\documentclass[11pt]{article}

\PassOptionsToPackage{table}{xcolor}
\usepackage[preprint]{acl}

\usepackage{times}
\usepackage{latexsym}

\usepackage[T1]{fontenc}

\usepackage[utf8]{inputenc}

\usepackage{microtype}

\usepackage{inconsolata}

\usepackage{graphicx}

\usepackage{hyperref}

\usepackage{url}
\usepackage{graphicx}
\usepackage{booktabs}
\usepackage{multirow}
\usepackage{amsfonts}
\usepackage{amsmath}
\usepackage{nicefrac}
\usepackage{enumitem}
\usepackage{tcolorbox}
\usepackage{xcolor}
\usepackage[normalem]{ulem}

\newcommand{\score}[2]{#1 {\scriptsize$\pm #2$}}
\newcommand{\tblhead}[1]{\textcolor{black}{\textbf{#1}}}

\tcbuselibrary{breakable,listings,skins}
\ifXeTeX
\else
  \usepackage{microtype}
\fi
\setlist[itemize,enumerate]{topsep=2pt,itemsep=0pt,parsep=0pt,partopsep=0pt}

\newtcblisting{promptbox}[1]{
  breakable,
  colback=teal!5,
  colframe=teal!40,
  coltitle=teal!70!black,
  title=\textbf{#1},
  fonttitle=\bfseries,
  boxrule=0.6pt,
  arc=2mm,
  left=1mm,
  right=1mm,
  top=1mm,
  bottom=1mm,
  listing only,
  listing options={
    basicstyle=\footnotesize\ttfamily,
    breaklines=true,
    breakatwhitespace=true
  }
}
\usepackage{relsize}
\usepackage{booktabs}

\usepackage{silence}
\title{Feedback-to-Rubrics: Can We Learn Expert Criteria \\ from Inline Comments?}

\author{
\textbf{Kotaro Yoshida\textsuperscript{1,2,*}},\quad
\textbf{So Kuroki\textsuperscript{1,*}},\quad
\textbf{Yuki Imajuku\textsuperscript{1}},\quad
\textbf{Taishi Nakamura\textsuperscript{1,2}},\\
\textbf{Ryunosuke Iwai\textsuperscript{1}},\quad
\textbf{Haruki Goda\textsuperscript{1}},\quad
\textbf{Takuya Akiba\textsuperscript{1}},\\[0.5em]
\textsuperscript{1}Sakana AI,\quad
\textsuperscript{2}Institute of Science Tokyo \\
\small{\texttt{\{kotaroyoshida,sokuroki\}@sakana.ai}}
}

\begin{document}
\maketitle

{
  \renewcommand{\thefootnote}{\fnsymbol{footnote}}
  \footnotetext[1]{Equal contribution.}
}

\begin{abstract}
Large language models (LLMs) are increasingly used for writing and review support, but their usefulness depends on context-dependent criteria, such as expert preferences or organization-specific conventions, that are often tacit, undocumented, and difficult to elicit directly.
We propose a problem setting for learning reusable natural-language rubrics from accumulated inline comments on artifacts such as human-written or LLM-generated drafts.
Our method infers rubrics from these comments and iteratively refines them by observing comment-wise mismatches between rubric-conditioned predictions and reference comments.
We evaluate the proposed method in real-world review settings and in controlled settings with reference rubrics. These results show that inline comments can be distilled into reusable rubrics that support comment prediction, rubric understanding, and automatic artifact revision.
\end{abstract}

\begin{figure*}[t!]
  \centering
  \vspace{-3mm}
  \includegraphics[width=0.95\textwidth]{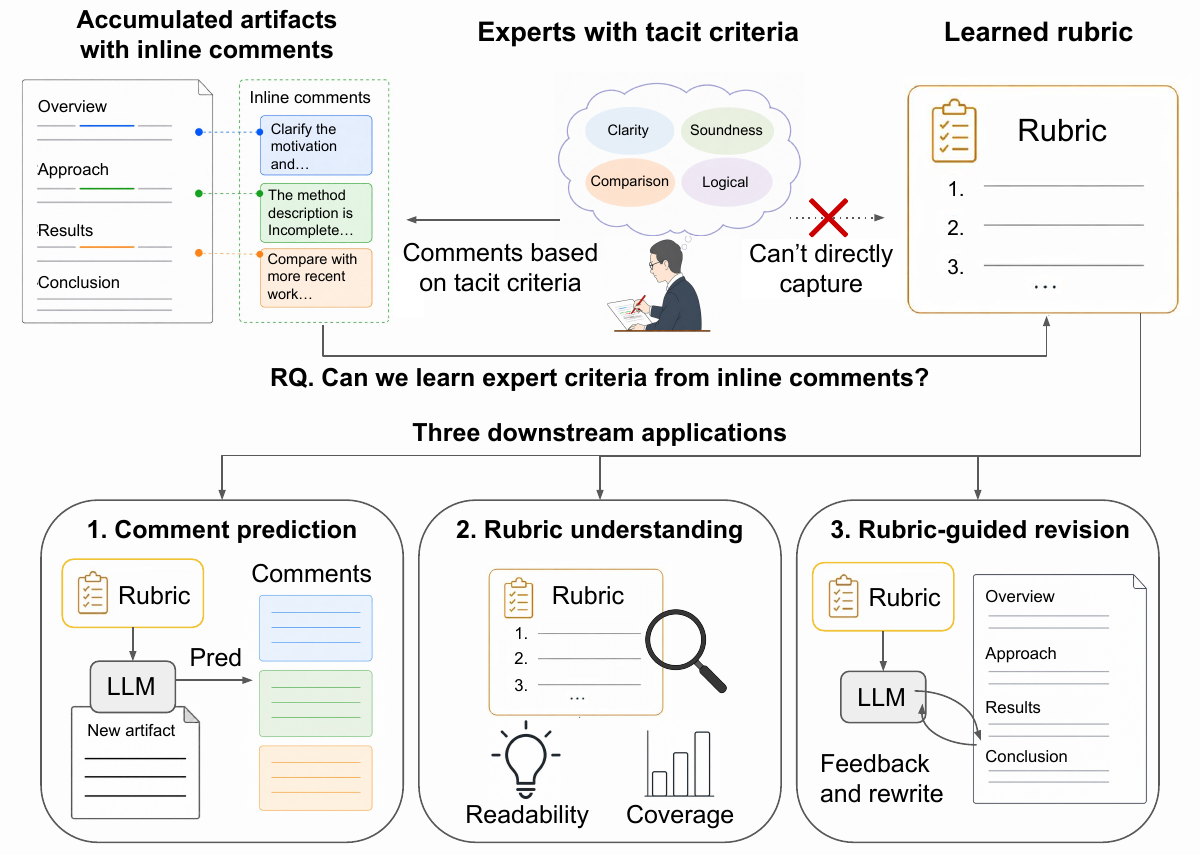}
  \vspace{-3mm}
  \caption{
\textbf{Conceptual overview of Feedback-to-Rubrics and downstream applications.}
\textbf{Top:} Feedback-to-Rubrics assumes a collection of artifacts with experts' inline comments and poses the main research question (RQ): \textit{Can we learn expert criteria from inline comments?} The method treats these comments as observations of tacit expert criteria that are difficult to elicit directly and uses them to learn an explicit rubric. \textbf{Bottom:} The learned rubric is then used for three downstream tasks: (1) comment prediction, (2) rubric understanding, and (3) rubric-guided artifact revision.}
  \label{fig:concept}
  \vspace{-3mm}
\end{figure*}

\section{Introduction}

Large language models (LLMs) are increasingly used to support writing and review~\citep{lee2024design}, but their usefulness is limited when they lack context-dependent criteria, such as expert preferences or organization-specific conventions.
A rubric makes such criteria explicit and guides LLM outputs toward expert judgments.
In practice, however, the criteria that matter in a given application are often tacit, undocumented, and difficult to elicit directly from humans.

This difficulty has motivated work that learns such criteria from observed human judgments.
For example, recent studies have inferred interpretable natural-language criteria from human behavior or preference data and evaluated them through held-out judgment prediction or reconstruction \citep{lin-etal-2024-interpretable, findeis-etal-2025-icai}.
At the same time, studies on explicit rubrics have shown that they can improve LLM evaluation and feedback generation \citep{hashemi-etal-2024-llm,yuan-etal-2024-llmcrit,kim2024prometheus}.
Together, these lines of work suggest a promising route: infer natural-language criteria from observed human behavior, and use those criteria as reusable conditioning signals for LLMs.

However, these approaches to rubric learning and rubric use are not directly applicable to review settings, where feedback is often provided as inline comments.
Existing rubric learning studies assume relatively global judgment signals, such as satisfaction behavior over an entire dialogue or pairwise preferences between candidate responses \citep{lin-etal-2024-interpretable,findeis-etal-2025-icai}. 
In contrast, inline comments are strongly tied to local contexts and reveal which concern a reviewer raises for a local text span and at what level of specificity \citep{hanawa-etal-2021-exploring,nagata-2019-toward}.
Therefore, learning a rubric from inline comments requires identifying and organizing the local criteria that govern comment behavior.
Although prior work on feedback has studied inline comment generation and span localization, these approaches do not focus on rubric learning \citep{nagata-2019-toward,hanawa-etal-2021-exploring,hellman-etal-2020-multiple}.

To address this gap, we propose \emph{Feedback-to-Rubrics}, a new problem setting that formalizes rubric learning from inline comments. 
Figure~\ref{fig:concept} (Top) illustrates this setting: expert inline comments on text artifacts, such as human-written or LLM-generated drafts, are treated as observations of tacit criteria, and the goal is to organize those criteria into an explicit, reusable rubric.
This setting is challenging because it requires learning a reusable rubric whose criteria remain grounded in local, span-level feedback while generalizing beyond individual comments.

We tackle this challenge by proposing a comment-prediction-based method that learns rubrics from inline comments and refines their criteria using local comment mismatches.
Our method infers an initial rubric from accumulated inline comments, uses it to predict comments, and iteratively refines the rubric by comparing the predicted comments with reference comments and updating the rubric based on the differences.
To capture local information, we introduce a comment-wise refinement signal that links each comment-level mismatch to the criteria used to generate it, enabling updates to the granularity of individual criteria.

We evaluate the learned rubric across nine diverse tasks in the three downstream applications shown in Figure~\ref{fig:concept} (Bottom): (1) comment prediction, where LLMs conditioned on refined rubrics make more accurate comment predictions for held-out artifacts than no-rubric and retrieval-based baselines \citep{nagata-2019-toward,lewis2020retrieval}; (2) rubric understanding, where we qualitatively inspect the natural-language rubrics and quantitatively evaluate them against reference rubrics in terms of coverage and precision trade-offs; and (3) rubric-guided artifact revision, where LLMs conditioned on refined rubrics improve iterative revision over both no-rubric and initial-rubric conditions. 
Together, these evaluations show that rubrics learned from inline comments are useful as reusable conditioning signals across diverse tasks beyond comment prediction.

The contributions of this paper are as follows.
\begin{itemize}[leftmargin=11pt]
  \item We introduce Feedback-to-Rubrics, a problem setting for learning reusable natural-language rubrics from accumulated inline comments.
  \item We propose a comment-prediction-based method that learns rubrics from inline comments and refines them using local comment mismatches.
  \item We evaluate the learned rubrics in three applications: comment prediction, rubric understanding, and rubric-guided artifact revision.
\end{itemize}

\section{Related Work}

This work lies at the intersection of research that treats inline comments as generation targets and research that externalizes human judgment criteria as rubrics.

\subsection{Inline Comment Prediction and Analysis}

Prior work has studied inline comments both as generation targets and as observations of evaluation behavior. Studies on feedback comment generation model local feedback behavior, including comment prediction and span localization \citep{nagata-2019-toward,nagata2021shared,hanawa-etal-2021-exploring,hellman-etal-2020-multiple}. These studies show that feedback can be formulated around local spans or errors, rather than only as holistic artifact-level assessment. Studies on peer-review analysis instead treat comments as observations of evaluation behavior. In this domain, PeerRead introduced a large-scale peer-review dataset \citep{kang-etal-2018-dataset}, and subsequent work has analyzed dimensions of review comments, such as substantiation, utility, and review aspect \citep{guo-etal-2023-automatic,sadallah-etal-2025-good,lu-etal-2025-identifying}. Other work links comments to revision edits, showing that comments can also act as interventions in revision behavior \citep{darcy-etal-2024-aries}.

In contrast, we formulate Feedback-to-Rubrics as a new problem setting for learning reusable evaluation criteria from accumulated inline comments.
This differs from prior work that treats comments as generation targets or objects of analysis: we use comment prediction as a testbed for learning and refining an interpretable rubric that explains expert comment behavior.

\subsection{Rubric Learning for LLMs}

Prior work has used rubrics as explicit intermediate representations for diagnostic evaluation and feedback, including educational assessment and automated scoring \citep{fiacco-etal-2023-towards,eltanbouly-etal-2025-trates}. These studies show that rubrics can make evaluation criteria more explicit and interpretable. Recent LLM studies further show that rubrics can be used, generated, and applied as flexible representations for controlling evaluation and feedback generation \citep{hashemi-etal-2024-llm,yuan-etal-2024-llmcrit,fan-etal-2024-sedareval,gupta-etal-2025-carmo,siro-etal-2026-learning}.

The work most closely related to ours learns explicit judgment criteria from observed human behavior or preferences. For example, one study learns interpretable satisfaction rubrics from observed dialogue behavior and validates them through held-out prediction \citep{lin-etal-2024-interpretable}. Another study compresses pairwise preference data into a constitution and evaluates it through annotation reconstruction \citep{findeis-etal-2025-icai}. These studies show that human judgments can be externalized as natural-language intermediate representations and used to predict or control future evaluation behavior.

However, many rubric learning studies rely on global judgment signals. In contrast, our method learns from locally grounded inline comments and uses comment-level matches as signals for refining the rubric.

\section{Preliminaries: Inline Comment Prediction}
\label{sec:preliminaries}

This section defines the inline comment prediction setting and the evaluation protocol used by our method.
Section~\ref{sec:method} describes how we infer and update rubrics from inline comments.

Our data consist of multiple artifacts, each with multiple inline comments.
Let $x_i$ denote an artifact, $q_{ij}$ denote the $j$-th target quote (a text span) in that artifact, and $y_{ij}$ denote the corresponding reference comment.
A reference comment is typically a comment written by a human reviewer, but it can also be a task-provided target comment.
If multiple reference comments are associated with the same span, we treat each $(q_{ij}, y_{ij})$ pair as a separate instance by assigning a distinct index $j$.
We write the full dataset as
\[
  \mathcal{D}=\{(x_i,\{(q_{ij},y_{ij})\}_{j=1}^{m_i})\}_{i=1}^{N}.
\]
Appendix~\ref{app:representative-example} shows examples of artifact excerpts, target quotes, and reference comments.

Following prior feedback comment generation settings \citep{nagata-2019-toward,hanawa-etal-2021-exploring}, we assume that the target quote $q_{ij}$ is given and focus on generating a comment that matches the paired reference comment $y_{ij}$.
We split the dataset $\mathcal{D}$ into train, validation, and test sets.
The train split is used for learning, the validation split for selection, and the test split only for final evaluation.

Our evaluation measures how well generated comments align with reference comments.
For each split, we compute the mean content score over an evaluation set $\mathcal{T}$ as $\frac{1}{|\mathcal{T}|}\sum_{(i,j)\in\mathcal{T}} s_{ij}$.
Prior work on feedback comment generation and text generation evaluation often uses BLEU, ROUGE, or BERTScore to measure comment similarity \citep{nagata-2019-toward,hanawa-etal-2021-exploring,papineni-etal-2002-bleu,lin-2004-rouge,zhang-etal-2020-bertscore}.
However, inline comments make this comparison harder.
The same concern can be expressed in different words, and two comments can look similar on the surface or in embedding space while differing in concern, specificity, or actionability \citep{sadallah-etal-2025-good}.
Recent work suggests that generative LLMs can better capture domain-specific semantic agreement than encoder-only similarity models \citep{gatto2023text}.

Motivated by this finding, we use an LLM judge to compare the generated comment $\hat{y}_{ij}$ and the reference comment $y_{ij}$, conditioned on the target quote $q_{ij}$ and the artifact $x_i$.
The judge returns a content score $s_{ij}$ that measures whether $\hat{y}_{ij}$ captures the same concern, granularity, and direction of explanation as $y_{ij}$.
Because the target quote and reference comment are fixed, the judge makes a local, reference-guided semantic comparison rather than a holistic artifact-quality judgment.
This also follows recent reference-guided, rubric-based
evaluation and pairwise-comparison-based LLM evaluator
studies showing strong alignment with human judgments
in fine-grained text evaluation settings \citep{kim2024prometheus,liu2024aligning}.

\begin{figure*}[t]
  \centering
  \includegraphics[width=0.98\textwidth]{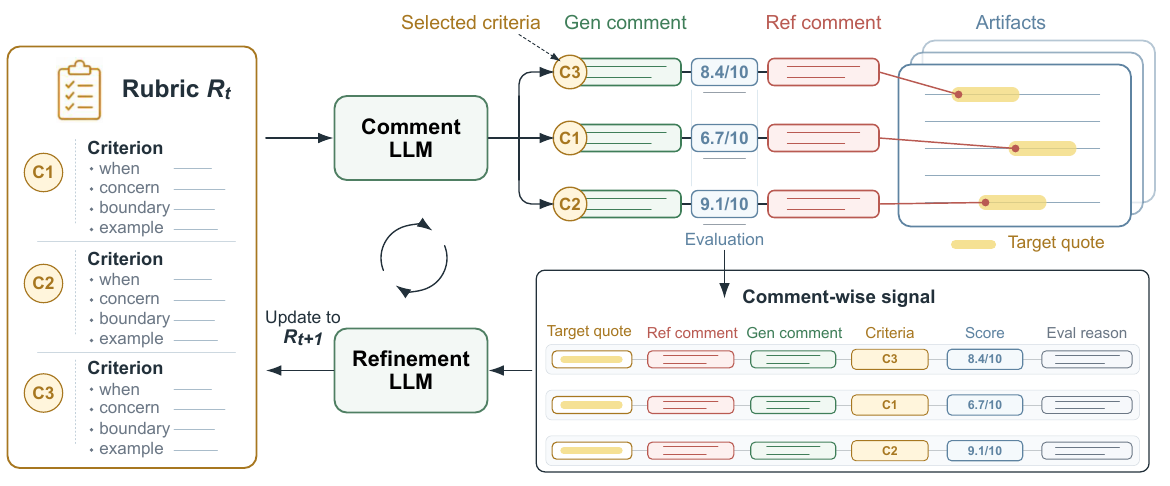}
  \vspace{-1em}
  \caption{\textbf{Method overview.} Starting from an initial rubric $R_0$ inferred from inline comments, the upper part predicts comments for target quotes conditioned on the current rubric $R_t$ and evaluates them against reference comments. The lower part then refines the rubric into $R_{t+1}$ using comment-wise signals from these prediction/evaluation results.}
  \label{fig:method-overview}
\end{figure*}

\section{Method}
\label{sec:method}

This work introduces a new problem, Feedback-to-Rubrics, which aims to learn the underlying criteria used by experts from inline comments.
As shown in Figure~\ref{fig:method-overview}, our method represents the learned criteria as a rubric and improves this rubric through an iterative loop.
This loop has two components: rubric-conditioned comment prediction and evaluation, which checks how well the rubric explains local comment behavior, and comment-wise rubric refinement, which updates the rubric based on those results.
Appendix~\ref{app:evaluation-prompts} lists the prompts used in our method.

\subsection{Feedback-to-Rubrics}

Feedback-to-Rubrics formalizes the problem of learning a reusable rubric from accumulated inline comments.
Given the dataset $\mathcal{D}$ defined in Section~\ref{sec:preliminaries}, whose instances consist of artifacts $x_i$, target quotes $q_{ij}$, and reference comments $y_{ij}$, the goal is to infer explicit criteria that explain what concerns experts raise, when they raise them, and at what level of granularity.
In this work, a rubric at round $t$ is defined as a set of criteria $R_t = \{ c_k^{(t)} \}_{k=1}^{K_t}$, where $K_t$ denotes the number of criteria at round $t$.
Each criterion $c_k$ is written in natural language and contains at least four elements: (1) applicability conditions that specify when to comment in a local context, (2) the criterion text that specifies which concern to state and at what granularity, (3) boundaries that prevent confusion with nearby criteria, and (4) examples of representative target quotes and reference comments.
The rubric makes explicit which concerns should be raised under which local conditions and at what granularity.
As a natural-language intermediate representation of expert criteria, the rubric supports downstream uses such as rubric understanding, rubric-conditioned comment prediction, and rubric-guided artifact revision.

To solve this problem, our method first infers an initial rubric $R_0$ from the training split $\mathcal{D}_{\mathrm{train}}\subset\mathcal{D}$ so that it explains the observed comment behavior.
This $R_0$ is not assumed to be correct; it is an initial hypothesis about the latent judgment criteria expressed through the comments.
It gives the method an explicit starting point that can be tested through comment prediction and revised through refinement.

\subsection{Comment Prediction and Evaluation}

Given a candidate rubric, comment prediction and evaluation test whether the rubric can reproduce expert comment behavior on target quotes.
In this setting, a comment LLM $G$ takes an artifact $x_i$, a target quote $q_{ij}$, and a rubric $R_t$ as input, and generates a natural-language comment for the quote.
Because the next step must know which criteria led to each generated comment, we ask the comment LLM to output both the comment text and the criteria it uses during generation.
We write the prediction at round $t$ as
\[
  (\hat{y}_{ij}^{(t)}, C_{ij}^{(t)}) = G(x_i, q_{ij}, R_t),
\]
where $C_{ij}^{(t)} \subseteq R_t$ denotes the subset of criteria used during generation for the $(i,j)$-th instance.

After generation, evaluation checks whether the rubric-conditioned comment $\hat{y}_{ij}^{(t)}$ reproduces the reference comment $y_{ij}$ and records why it succeeds or fails.
An evaluation LLM $J$ returns a content score $s_{ij}^{(t)}$ and judge reasoning $e_{ij}^{(t)}$ for each target quote:
\[
  (s_{ij}^{(t)}, e_{ij}^{(t)}) = J(x_i, q_{ij}, y_{ij}, \hat{y}_{ij}^{(t)}).
\]
The main metric is the reproducibility of comment content.
The evaluation emphasizes whether the generated comment captures the same concern, matches the reference granularity, and avoids adding unrelated critiques.
Because surface overlap and embedding similarity alone cannot reliably support this judgment, we use a context-aware evaluation LLM and use its reasoning as a diagnostic explanation for refinement.
Thus, prediction and evaluation turn each target quote into both a performance measurement and an update signal for the rubric.

\subsection{Comment-Wise Iterative Rubric Refinement}

Iterative rubric refinement is the update step in our proposed method: it converts prediction failures into concrete changes to the current rubric.
A refinement LLM updates $R_t$ so that the next rubric better explains the reference-comment behavior.
The refinement LLM must distinguish among overly broad criteria, missing concerns, unclear applicability conditions, redundant criteria, and criteria that already align well with the reference comments.
To support this decision, we provide the refinement LLM with not only the final score, but also the generated comment, the corresponding reference comment, the judge reasoning, and the criteria selected during generation.

However, if these signals are given only as separate field-wise lists, the link between a criterion and the specific comment in which it helped or failed is lost.
We therefore organize them as a comment-wise refinement signal, which keeps each prediction result paired with the criteria used to generate it.
For each instance at round $t$, we define the refinement signal as
\[
  h_{ij}^{(t)} =
  \bigl(q_{ij}, y_{ij}, \hat{y}_{ij}^{(t)}, s_{ij}^{(t)}, e_{ij}^{(t)}, C_{ij}^{(t)}\bigr),
  \mathcal{H}_t = \{h_{ij}^{(t)}\}.
\]

Given the current rubric $R_t$ and the comment-wise refinement signals $\mathcal{H}_t$, the refinement LLM generates the next rubric $R_{t+1}$.
By presenting the prediction result together with the selected criteria for each comment instance, the refinement LLM can update the rubric while distinguishing local successes and failures for each criterion.
Specifically, it can add a new criterion when a concern in the reference comment is missing, split or rewrite an overly broad criterion into more specific conditions, clarify the applicability conditions, boundaries, and examples of incorrectly applied criteria, and merge or remove redundant criteria.
It preserves criteria that already align well with the reference comments.
The updated rubric is then used in the next prediction/evaluation round, so each iteration tests the current rubric and revises it based on observed failures.

\section{Experiments}

\begin{table*}[t]
  \caption{\textbf{Ablation and baseline comparison on comment prediction.}
  We report mean content scores $s_{ij} \in [0,10]$ (higher is better) averaged over five runs, with standard deviations. Our full method, which uses the comment-wise refinement signal, outperforms all baselines on average.}
  \label{tab:ablation_expertlong}
  \centering
  \begingroup
  \normalsize
  \setlength{\tabcolsep}{2pt}
  \renewcommand{\arraystretch}{1.05}
  \resizebox{\textwidth}{!}{%
  \begin{tabular}{l c c c c c c c c c c}
    \toprule
    \tblhead{Method} &
    \begin{tabular}[c]{@{}c@{}}\tblhead{Research}\\\tblhead{proposal}\\\tblhead{review}\end{tabular} &
    \begin{tabular}[c]{@{}c@{}}\tblhead{Essay}\\\tblhead{review}\end{tabular} &
    \begin{tabular}[c]{@{}c@{}}\tblhead{Medical chat}\\\tblhead{annotation}\end{tabular} &
    \tblhead{Bio} &
    \tblhead{Chemical} &
    \tblhead{Cyber} &
    \tblhead{Edu} &
    \tblhead{Health} &
    \tblhead{Material} &
    \tblhead{Avg.} \\
    \midrule

    \multicolumn{11}{@{}l}{\textit{Retrieval-based methods}} \\
    \begin{tabular}[c]{@{}l@{}}Top-1 retrieval~\citep{nagata-2019-toward}\end{tabular} &
    \score{0.71}{0.07} &
    \score{3.01}{0.07} &
    \score{0.20}{0.09} &
    \score{1.68}{0.07} &
    \score{1.56}{0.11} &
    \score{3.34}{0.10} &
    \score{2.05}{0.09} &
    \score{1.39}{0.07} &
    \score{0.28}{0.11} &
    $1.58$ \\

    \begin{tabular}[c]{@{}l@{}}Top-3 RAG with LLM~\citep{lewis2020retrieval}\end{tabular} &
    \score{3.05}{0.46} &
    \score{8.39}{0.09} &
    \score{7.08}{0.28} &
    \score{2.30}{0.28} &
    \score{3.12}{0.18} &
    \score{4.76}{0.22} &
    \score{2.26}{0.26} &
    \score{2.75}{0.18} &
    \score{2.06}{0.24} &
    $3.97$ \\

    \midrule
    \multicolumn{11}{@{}l}{\textit{Rubric-based methods (Component ablation)}} \\
    \rowcolor{gray!10}
    No rubric &
    \score{3.00}{0.57} &
    \score{7.98}{0.12} &
    \score{6.62}{0.19} &
    \score{0.56}{0.14} &
    \score{1.76}{0.20} &
    \score{1.18}{0.23} &
    \score{1.37}{0.13} &
    \score{1.82}{0.35} &
    \score{1.79}{0.35} &
    $2.90$ \\

    \rowcolor{gray!10}
    + Initial rubric &
    \score{3.21}{0.62} &
    \score{8.54}{0.30} &
    \score{6.72}{0.41} &
    \score{2.66}{0.40} &
    \score{3.42}{0.36} &
    \score{7.13}{0.30} &
    \score{3.96}{0.36} &
    \score{3.84}{0.45} &
    \textbf{\score{2.43}{0.36}} &
    $4.66$ \\

    \rowcolor{gray!10}
    + Iterative rubric refinement &
    \score{3.32}{0.73} &
    \score{8.60}{0.03} &
    \score{7.15}{0.17} &
    \score{2.61}{0.61} &
    \score{3.64}{0.38} &
    \score{7.00}{0.23} &
    \textbf{\score{4.11}{0.41}} &
    \score{4.21}{0.48} &
    \score{2.35}{0.45} &
    $4.78$ \\

    \rowcolor{gray!10}
    + Comment-wise refinement signal &
    \textbf{\score{3.42}{0.30}} &
    \textbf{\score{8.71}{0.07}} &
    \textbf{\score{7.31}{0.21}} &
    \textbf{\score{2.75}{0.48}} &
    \textbf{\score{3.73}{0.38}} &
    \textbf{\score{7.43}{0.20}} &
    \score{3.95}{0.34} &
    \textbf{\score{4.75}{0.23}} &
    \score{2.31}{0.40} &
    $\mathbf{4.93}$ \\

    \bottomrule
  \end{tabular}%
  }
  \endgroup
\end{table*}

We evaluate Feedback-to-Rubrics along two axes: whether our method can learn reusable rubrics from accumulated inline comments, and whether the learned rubrics are useful for downstream applications.
We test these axes through three uses introduced in Figure~\ref{fig:concept}: comment prediction, rubric understanding, and rubric-guided artifact revision.
\begin{itemize}[leftmargin=11pt]
  \item \textbf{Q1:} Does the learned rubric improve comment prediction (Section~\ref{sec:comment-prediction-results})?
  \item \textbf{Q2:} Can the learned rubric serve as an interpretable intermediate representation, and does refinement improve the rubric itself (Section~\ref{sec:rubric-representation-results})?
  \item \textbf{Q3:} Does the learned rubric guide artifact revision for new artifacts (Section~\ref{sec:artifact_improvement})?
\end{itemize}

\subsection{Datasets and Experimental Setup}

\noindent\textbf{Datasets.}
We evaluate on nine tasks, including two real-world inline-comment tasks and rubric-grounded benchmark settings.
The real-world tasks test whether the method works on naturally accumulated feedback: research proposal review~\citep{zyska2026expos} and essay review~\citep{nagata2021shared}.
Publicly available inline-comment datasets suitable for research use are limited, and datasets that also include reference rubrics are particularly scarce.
We therefore construct additional benchmark settings from HealthBench~\citep{arora2025healthbench} and ExpertLongBench~\citep{ruan2025expertlongbench}, covering medical chat annotation and six ExpertLongBench domains: Bio, Chemical, Cyber, Edu, Health, and Material.
For these benchmark settings, we use an LLM to synthesize artifacts from the original task prompts and then create rubric-grounded target-quote/comment pairs.
Since ExpertLongBench provides domain-level global reference rubrics, we use it for rubric understanding and rubric-guided revision in addition to comment prediction.
Across all datasets, we split at the artifact level with a 0.6/0.2/0.2 ratio for train, validation, and test.
Appendix~\ref{app:datasets} provides construction details and split statistics.

\noindent\textbf{Experimental setup.}
The main-text results use Gemini 3.1 Pro preview.
We also run comment prediction with DeepSeek v4 Pro and GPT-5.4 low (see Appendix~\ref{app:additional-model-results}) on a three-domain subset.
To keep scoring comparable, all generated comments are judged by the same Gemini 3.1 Pro preview evaluator.
Predicted comments are scored with the 0--10 content score $s_{ij}$, and final test results report means and standard deviations over five runs.
Rubrics are learned from training comments, refined for 10 rounds using training-split prediction signals, selected by validation scores, and evaluated on the test split.
We call the validation-selected rubric from the full comment-wise refinement condition the Best-val rubric.
We compare rubric-conditioned generation against no-rubric generation.
For retrieval baselines, we use Top-1 retrieval, following prior feedback comment generation work~\citep{nagata-2019-toward}, and Top-3 RAG with LLM, following retrieval-augmented generation and in-context example selection~\citep{lewis2020retrieval,ram2023context,rubin-etal-2022-learning,liu-etal-2022-makes}.
Appendix~\ref{app:experimental-setup-details} provides refinement and baseline details.

\begin{figure*}[t]
  \centering
  \begin{minipage}[t]{0.24\textwidth}
    \vspace{0pt}
    \centering
    \includegraphics[width=\linewidth]{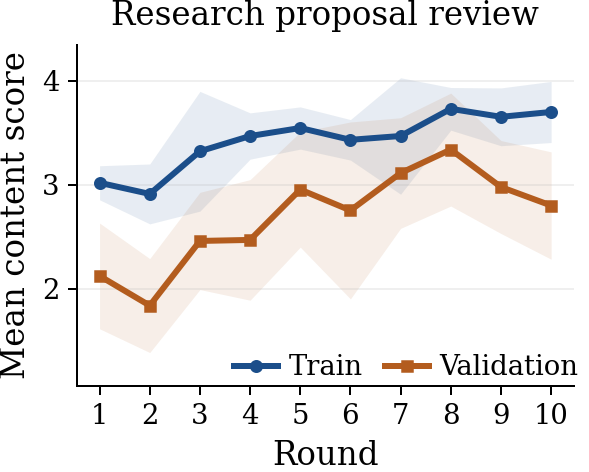}
  \end{minipage}
  \hfill
  \begin{minipage}[t]{0.24\textwidth}
    \vspace{0pt}
    \centering
    \includegraphics[width=\linewidth]{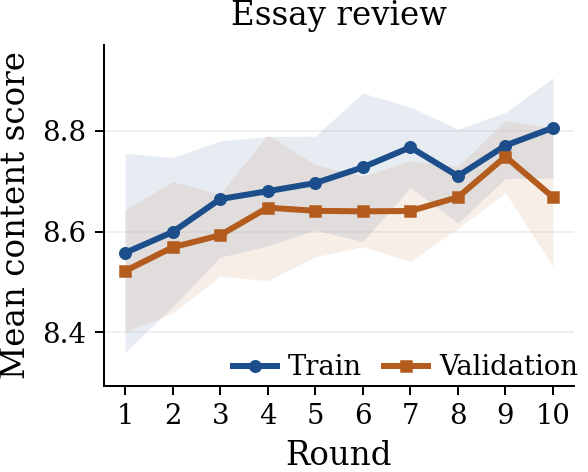}
  \end{minipage}
  \hfill
  \begin{minipage}[t]{0.24\textwidth}
    \vspace{0pt}
    \centering
    \includegraphics[width=\linewidth]{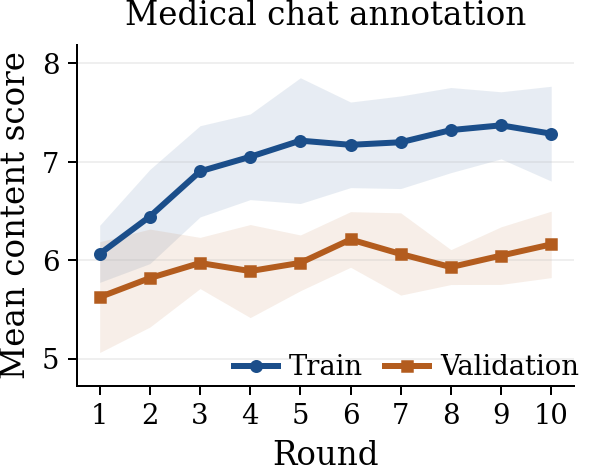}
  \end{minipage}
  \hfill
  \begin{minipage}[t]{0.24\textwidth}
    \vspace{0pt}
    \centering
    \includegraphics[width=\linewidth]{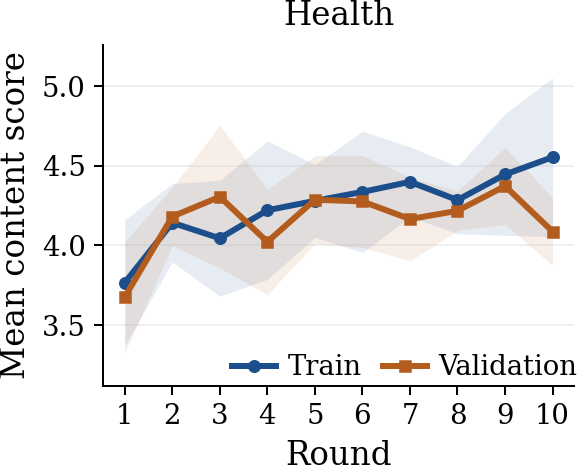}
  \end{minipage}
  \caption{\textbf{Train and validation performance across refinement rounds}. Each plot shows the mean content score on the train and validation splits. In ExpertLongBench, we show the results for Health here and defer the remaining domains to Appendix~\ref{app:other-score-curves}. In all tasks, both curves increase steadily over rounds, suggesting that refinement improves comment prediction without clear overfitting.}
  \label{fig:train-val-curves}
\end{figure*}

\subsection{Q1: Comment Prediction}
\label{sec:comment-prediction-results}

Q1 asks whether a learned rubric helps an LLM predict comments for new artifacts.

\noindent\textbf{Component ablation.}
Table~\ref{tab:ablation_expertlong} reports component-ablation rows that compare four ablation settings: no rubric, an initial learned rubric, iterative refinement using aggregate field-wise signals, and comment-wise refinement that keeps each prediction paired with the criteria used for that instance.
The average score increases from 2.90 without a rubric to 4.66 with an initial learned rubric, and further improves to 4.93 when comment-wise signals are used.
This average trend, together with improvements in most individual task columns, indicates that updating the rubric based on alignments between generated and reference comments helps capture task-specific comment behavior. 
As shown in Figure~\ref{fig:train-val-curves}, scores improve steadily across refinement rounds without clear overfitting.
The remaining ExpertLongBench curves are reported in Appendix~\ref{app:other-score-curves}.
Appendix~\ref{app:additional-model-results} further shows the same overall pattern with DeepSeek v4 Pro and GPT-5.4 low backends.

\noindent\textbf{Comparison with retrieval baselines.}
Table~\ref{tab:ablation_expertlong} also reports rows for retrieval-based baselines. 
Top-1 retrieval performs poorly on average (1.58), while Top-3 RAG improves to 3.97 but still falls below the initial rubric (4.66) and the full refinement setting (4.93).
This suggests that learning reusable criteria from feedback is more effective for comment generation than reusing nearby past comments.

\begin{figure*}[t]
\centering
\begin{tcolorbox}[
  colback=cyan!5,
  colframe=cyan!30,
  coltitle=cyan!50!black,
  title=\textbf{Example learned rubric items for research proposal review},
  fonttitle=\bfseries,
  boxrule=1pt,
  arc=2mm,
  left=0mm,
  right=0mm,
  top=0mm,
  bottom=0mm
]
{\small\textbf{Initial rubric} (5 items)}
\vspace{-4pt}
\begin{lstlisting}[basicstyle=\small]
[-] Factual claims, statistics, or references to existing systems or studies that lack citations or concrete examples.
[-] Informal, dramatic, storytelling, or marketing language in place of neutral scientific tone.
[-] Research questions that are overly broad, imprecise, or lack a specific and measurable scope.
[-] Theoretical frameworks that are named but not explained or connected to the proposed research.
[-] Methodological steps that lack concrete details such as metrics, sample sizes, or model choices.
\end{lstlisting}
\vspace{-4pt}
{\small\textbf{Best-val rubric} (9 items)}
\vspace{-4pt}
\begin{lstlisting}[basicstyle=\small]
[-] Factual claims or references to existing systems, studies, or frameworks without citations or concrete evidence.
[-] Informal, dramatic, storytelling, conversational, or marketing language in place of neutral scientific tone.
[-] Research questions that are overly broad, imprecise, lack measurable scope, or are disjointed without a unifying focus.
[-] Theoretical frameworks or foundational concepts named but not explained or connected to the research, method, or task.
[-] Methods or study designs lacking detail, feasibility justification, or alignment with research questions.
[-] Content misplaced within sections, prematurely introduced research questions, or placeholder titles.
[-] Wording that is unclear, ambiguous, redundant, or grammatically incorrect.
[-] Literature-review sections that list papers without synthesis, comparison, or critique.
[+] Well-formulated limitations, helpful footnotes, proper research-question handling, or well-described target groups.
\end{lstlisting}
\vspace{-6pt}
\end{tcolorbox}
\setlength{\abovecaptionskip}{0pt}
\caption{\textbf{Example learned rubric items for research proposal review.}
The leading ``[--]'' marks negative criteria, whereas ``[+]'' marks positive criteria.
Each line is a condensed summary of a rubric item; full texts are shown in Appendix~\ref{app:research-proposal-rubric-full}.}
\label{fig:rubric-compressed}
\end{figure*}

\subsection{Q2: Rubric Understanding}
\label{sec:rubric-representation-results}

\begin{table}[t]
\centering
\scriptsize
\caption{\textbf{Improvement from the initial rubric to the Best-val rubric on the six ExpertLongBench domains.} Each value shows Best-val minus Initial. Scores use a 0--10 scale. H-mean denotes the harmonic mean of Recall and Precision. Overall, the Best-val rubric improves recall on average, with a slight average drop in precision and a net gain in H-mean.}
\setlength{\tabcolsep}{6pt}
\renewcommand{\arraystretch}{.9}
\resizebox{0.45\textwidth}{!}{%
\begin{tabular}{lcc|c}
\toprule
\tblhead{Task}
& \tblhead{Recall}
& \tblhead{Precision}
& \tblhead{H-mean} \\
\midrule

Bio
& $\mathbf{+0.40}$
& $+0.00$
& $\mathbf{+0.22}$ \\

Chemical
& $\mathbf{+0.84}$
& $\mathbf{+1.04}$
& $\mathbf{+0.99}$ \\

Cyber
& $\mathbf{+0.08}$
& $-1.06$
& $-0.60$ \\

Edu
& $+0.00$
& $-0.72$
& $-0.46$ \\

Health
& $\mathbf{+3.80}$
& $-0.16$
& $\mathbf{+2.69}$ \\

Material
& $\mathbf{+1.86}$
& $-0.20$
& $\mathbf{+0.87}$ \\

\midrule
Avg.
& $\mathbf{+1.17}$
& $-0.19$
& $\mathbf{+0.62}$ \\

\bottomrule
\end{tabular}
}
\label{tab:rubric-compare-expertlongbench}
\end{table}

Q2 asks whether the learned rubric functions as an inspectable natural-language representation and whether refinement improves the quality of the rubric itself.

\noindent\textbf{Qualitative assessment.}
Figure~\ref{fig:rubric-compressed} presents an example from the research proposal review task, comparing summarized initial and Best-val rubric items. We use this task for qualitative inspection because it provides a familiar setting whose criteria are easy for the research community to interpret. Full rubric texts and annotated examples are provided in Appendix~\ref{app:research-proposal-rubric-full}.

The Best-val rubric covers broader and more fine-grained criteria than the initial rubric.
The initial rubric contains five items that already address core aspects of proposal quality, including missing citations, unscientific tone, overly broad research questions, unexplained theoretical frameworks, and underspecified methodology.
In contrast, the Best-val rubric expands to nine items, showing more specific selection conditions.
For example, refinement broadens existing criteria: citation support is extended from systems and studies to frameworks, and methodological detail is expanded to include feasibility and alignment with research questions.
It also introduces new criteria for structural organization, clarity and wording, literature synthesis, and positive feedback.
These observations indicate that the refinement process not only expands the coverage of criteria but also yields a rubric that captures finer-grained patterns of expert feedback that are absent from the initial rubric.

\noindent\textbf{Quantitative assessment.}
We quantitatively evaluate the agreement between the learned rubrics and the reference rubrics on the six ExpertLongBench domains.
We show the learned rubric and the reference rubric to an LLM judge and ask it to score recall and precision on a 0--10 scale.
Recall measures how well the learned rubric covers important criteria in the reference global rubric.
Precision measures how well the learned rubric avoids excessive or inappropriate criteria.
Appendix~\ref{app:rubric-prediction-prompt} shows the prompt used for rubric judging.

Table~\ref{tab:rubric-compare-expertlongbench} shows the results.
Compared with the initial rubric, the Best-val rubric does not reduce recall in any domain and increases it in five of the six domains, while precision drops slightly on average.
This suggests that iterative rubric refinement primarily improves coverage of reference criteria, with a precision trade-off in some domains; overall agreement still improves in four of the six domains.
Therefore, the improvement in comment prediction is not merely due to surface-level tuning.
Instead, it is consistent with a shift toward broader coverage of human judgment criteria, with a remaining precision trade-off.

\begin{table}[t]
\centering
\small
\caption{\textbf{Artifact revision results on the six ExpertLongBench tasks.} The table reports changes in the number of reference rubric items satisfied after three rounds of rubric-guided artifact revision. On average, providing an explicit rubric leads to larger gains than the no-rubric condition, and using the Best-val rubric yields the highest overall improvement.}
\setlength{\tabcolsep}{6pt}
\renewcommand{\arraystretch}{1.07}
\begin{tabular}{lccc}
\toprule
\tblhead{Task}
& \tblhead{No rubric}
& \tblhead{Initial}
& \tblhead{Best-val} \\
\midrule
Bio
& +\score{3.87}{0.11} & +\score{4.88}{0.18} & \textbf{+\score{4.95}{0.06}} \\
Chemical
& +\score{2.92}{0.15} & +\score{4.56}{0.30} & \textbf{+\score{4.73}{0.36}} \\
Cyber
& +\score{1.98}{0.34} & \textbf{+\score{2.10}{0.26}} & +\score{2.08}{0.20} \\
Edu
& \textbf{+\score{0.84}{0.16}} & +\score{0.82}{0.24} & +\score{0.78}{0.13} \\
Health
& +\score{6.29}{1.59} & +\score{6.86}{1.22} & \textbf{+\score{8.07}{1.27}} \\
Material
& +\score{4.53}{0.13} & +\score{4.97}{0.25} & \textbf{+\score{5.17}{0.20}} \\
\midrule
Avg.
& +3.41
& +4.03
& \textbf{+4.30} \\
\bottomrule
\end{tabular}
\label{tab:improvement-loop-results}
\end{table}

\subsection{Q3: Rubric-Guided Artifact Revision}
\label{sec:artifact_improvement}

Q3 asks whether a learned rubric can guide automatic artifact revision.
We compare three conditions: No rubric, Initial rubric, and Best-val rubric.
In the No rubric condition, the model revises the artifact without any explicit rubric.
In the rubric-based conditions (Initial and Best-val), the corresponding rubric is provided to the model, and the model is instructed to revise the artifact so that it satisfies the rubric criteria.
The initial artifacts are generated by an LLM to be low quality.
We evaluate the revised artifacts using the reference rubric provided for each ExpertLongBench domain.
Specifically, we assign one point for each reference rubric item that the revised artifact satisfies, and report the improvement in this score.
We set the total number of revision rounds to three.

The results are shown in Table~\ref{tab:improvement-loop-results}.
On average, both rubric-based conditions (Initial and Best-val) improve artifact revision over the No rubric condition.
In addition, when comparing the initial and Best-val rubrics, the Best-val rubric achieves a higher average improvement score.
These results suggest that the rubrics learned by our framework are useful for improving artifacts themselves in a way that is aligned with task-specific preferences.

\section{Conclusion}

This paper introduced Feedback-to-Rubrics, a problem setting for learning reusable rubrics from naturally accumulated inline feedback. We proposed a comment-prediction-based method that infers an initial rubric from inline comments and then refines it using comment-wise mismatches between generated and reference comments.
We evaluated the learned rubrics in three downstream uses: comment prediction, rubric understanding, and rubric-guided artifact revision. The learned rubrics improved comment prediction over no-rubric and retrieval-based baselines, covered reference criteria more broadly, and helped LLMs revise artifacts more effectively. These results show that inline feedback can be externalized into reusable, context-dependent criteria for evaluation, interpretation, and revision.

\section{Limitations}

This work has three main limitations.
First, our evaluation still relies on LLM-based evaluation of predicted comments. 
Although recent studies support reference-guided rubric-based evaluation and pairwise LLM evaluators \citep{kim2024prometheus,liu2024aligning}, agreement with human judgments in our specific setting should be further validated.
Second, following established feedback comment generation settings \citep{nagata-2019-toward,hanawa-etal-2021-exploring}, we assume that target quotes are given. 
This choice is also motivated by the fact that local one-to-one comparison between reference and generated comments supports more reliable LLM-based judgment, as discussed above, and provides a denser update signal for rubric refinement.
Extending Feedback-to-Rubrics to joint localization and commenting is future work.
Third, our rubric-understanding and artifact-revision evaluations rely on constructed benchmark settings. Inline comments naturally accumulate in education, research supervision, and workplace training, for example on student writing, research proposals, and internal documents. We expect Feedback-to-Rubrics to be useful for such privately accumulated feedback. However, such data are rarely public and even more rarely paired with reference rubrics. Therefore, in this work, we partially use rubric-grounded synthetic comments for proof-of-concept evaluation. Constructing research-ready datasets with naturally occurring comments and reference rubrics remains future work.

\bibliography{custom}

\clearpage
\appendix
\section*{Appendix}

\section{Datasets}
\label{app:datasets}

Table~\ref{tab:dataset-splits} reports statistics computed from the processed split files used in the experiments.
The split unit is always the artifact, so multiple comments attached to the same artifact never cross train, validation, or test boundaries.

\begin{table*}[t]
  \caption{\textbf{Dataset sources, split sizes, and aggregate statistics.} Comment counts are shown as train/validation/test, with corresponding artifact counts in parentheses. Lengths are average character counts over all splits.}
  \label{tab:dataset-splits}
  \centering
  \footnotesize
  \setlength{\tabcolsep}{1.5pt}
  \renewcommand{\arraystretch}{1.05}
  \begin{tabular}{@{}p{0.16\linewidth}p{0.22\linewidth}p{0.23\linewidth}p{0.07\linewidth}p{0.06\linewidth}p{0.07\linewidth}@{}}
    \toprule
    \tblhead{Data source} & \tblhead{Dataset} & \tblhead{Comments (Artifacts)} & \begin{tabular}[c]{@{}l@{}}\tblhead{Comm./}\\\tblhead{art.}\end{tabular} & \begin{tabular}[c]{@{}l@{}}\tblhead{Quote}\\\tblhead{len.}\end{tabular} & \begin{tabular}[c]{@{}l@{}}\tblhead{Comment}\\\tblhead{len.}\end{tabular} \\
    \midrule
    \begin{tabular}[t]{@{}l@{}}Exposia\\\citep{zyska2026expos}\end{tabular} & Research proposal review & 118/31/43 (7/2/3) & 16.00 & 115.1 & 98.1 \\
    \midrule
    \begin{tabular}[t]{@{}l@{}}Feedback Dataset\\\citep{nagata2021shared}\end{tabular} & Essay review & 329/110/113 (300/100/100) & 1.10 & 6.1 & 163.0 \\
    \midrule
    \begin{tabular}[t]{@{}l@{}}HealthBench\\\citep{arora2025healthbench}\end{tabular} & Medical chat annotation & 180/69/92 (30/10/10) & 6.82 & 153.7 & 323.3 \\
    \midrule
    \multirow[t]{6}{=}{\begin{tabular}[t]{@{}l@{}}ExpertLongBench\\\citep{ruan2025expertlongbench}\end{tabular}} & Bio & 214/74/75 (60/20/20) & 3.63 & 177.8 & 210.4 \\
     & Chemical & 177/58/58 (60/20/20) & 2.93 & 176.4 & 188.7 \\
     & Cyber & 209/62/71 (60/20/20) & 3.42 & 108.3 & 192.6 \\
     & Edu & 235/86/82 (60/20/20) & 4.03 & 91.8 & 163.6 \\
     & Health & 239/81/69 (60/20/20) & 3.89 & 94.7 & 153.6 \\
     & Material & 64/18/17 (30/10/10) & 1.98 & 199.5 & 283.7 \\
    \bottomrule
  \end{tabular}
\end{table*}

\subsection{Dataset Construction Details}

Across all datasets, we split at the artifact level with a 0.6/0.2/0.2 target split ratio for train, validation, and test.
The table reports the resulting split counts after this artifact-level split.

\noindent\textbf{Real-world inline-comment datasets.}
The processed research and essay data provide artifact-level inputs, target quotes, reference comments, and optional character offsets, but they do not expose per-comment annotator IDs in a uniform way.
For research proposal review, the selected Exposía subset is restricted to a single reviewer.
These two datasets are used as naturally occurring inline-comment settings and do not provide reference rubrics for rubric-level evaluation.

\noindent\textbf{Rubric-grounded benchmark settings.}
HealthBench and ExpertLongBench provide rubric information that allows us to construct benchmark settings with target quotes and reference comments.
HealthBench provides prompt-specific rubrics for evaluating medical responses, whereas ExpertLongBench provides domain-level global rubrics for expert-level long-form generation tasks.
For this reason, rubric-level agreement and rubric-guided revision are evaluated on ExpertLongBench, whose global reference rubrics match the single shared rubric setting used by our method.
For medical chat annotation, each synthetic reference comment is linked to the HealthBench criterion from which it was derived, including the criterion polarity.
For each prompt, we first sample a subset of HealthBench criteria and construct synthetic artifacts designed to satisfy half of the sampled criteria while violating the remaining half.
We then generate reference comments only for the violated criteria.
For these constructed benchmark settings, we report the available processed fields and aggregate statistics used in the experiments.
All input data had already been filtered before the experiments.
We did not apply additional empty-comment filtering or quote deduplication during the reported runs.
For ExpertLongBench, we construct feedback-style data from the publicly released benchmark.
We use the six domains reported in Table~\ref{tab:dataset-splits}: Bio, Chemical, Cyber, Edu, Health, and Material.
ExpertLongBench provides expert-level long-form generation tasks and reference global rubrics, but not quote-level feedback comments.
We therefore generate task outputs using the original task prompts, task-specific system prompts, and supplementary materials, and then generate quote-level reference comments grounded in the corresponding reference rubrics.
Both task-output generation and reference-comment generation use \texttt{google/gemini-3-flash-preview} with medium reasoning effort and temperature 1.0.
We increase output-length limits relative to the original configuration to reduce truncation.
In both benchmark sources, the constructed learning instances consist of artifacts, target quotes, and reference comments.
The source rubrics are used to construct benchmark instances and, for ExpertLongBench, to evaluate rubric-level agreement and rubric-guided revision; they are not used as learned attributes in the Feedback-to-Rubrics input.

Each processed input row is treated as one artifact and contains an artifact identifier, the original prompt when available, the artifact text to be reviewed, and the set of reference comments attached to the artifact.
Each reference comment contains a target quote, comment text, and optional character offsets.
Message-list prompts are normalized into role-and-content strings before being passed to the LLMs.
When source comments include existing rubric IDs, we do not use them as learned attributes.
Criterion IDs are used only as runtime traces of which learned criterion the generator cited.

Generation prompts include the full artifact, all rubric criteria, the target quote, and available start and end offsets.
The generation LLM returns one structured comment and cited criterion IDs for each target quote.
The implementation aligns returned IDs with the reference-comment order, stores valid criterion IDs as traces, and inserts a fallback comment if the generator omits a requested target quote.

\paragraph{Dataset Licenses.}
We used four datasets under their respective license terms. HealthBench is released under the MIT License. ExpertLongBench is licensed under CC BY-NC-SA 4.0. Exposía is licensed under CC BY-NC 4.0 unless otherwise noted. The Essay dataset does not specify a formal license, but its distribution terms permit use primarily for non-commercial research purposes. We used all datasets solely for research purposes and complied with the conditions stated on their respective dataset pages.

\section{Experimental Setup Details}
\label{app:experimental-setup-details}

\subsection{Main Protocol}
Comment prediction follows the fixed-target-quote setting in Section~\ref{sec:preliminaries}: the model receives an artifact, a target quote, and, when applicable, a rubric, and generates one comment for that quote.
Generated comments are evaluated with the content score $s_{ij}\in[0,10]$ using the prompt in Appendix~\ref{app:comment-evaluation-prompt}.
For each dataset, iterative rubric refinement runs for 10 rounds.
Training-split reference comments are used for refinement, the validation split is used for round selection, and the test split is used only for final evaluation.
During refinement, the model receives the current refinement signal and histories from up to three previous rounds.
We run each experimental condition five times on the same split and report the mean and standard deviation over these runs.

\subsection{Retrieval-Based Methods}
We compare our method with the following two retrieval-based baselines.

\begin{description}
\item[Top-1 retrieval]
For each test position, we encode the target quote as a query and retrieve the most similar comment from the training set.
The retrieved comment text is then copied verbatim as the prediction.
This baseline is purely retrieval-based: it does not call an LLM and does not use any learned rubric.
This baseline follows the retrieval setting used in prior feedback comment generation work~\citep{nagata-2019-toward}.

\item[Top-3 RAG with LLM]
For each test position, we retrieve the top $k=3$ most similar comments from the training set and provide them to an LLM as in-context examples.
This baseline follows retrieval-augmented generation and in-context example selection, where retrieved documents or similar examples are placed in the context of a frozen or API LLM~\citep{lewis2020retrieval,ram2023context,rubin-etal-2022-learning,liu-etal-2022-makes}.
The retrieved examples are included in the user prompt using the following format.
\begin{promptbox}{In-context Example Format}
### Position [position index]
  target_quote: "[target quote]"
    1. retrieved comment: "[retrieved comment 1]", target_quote: "[retrieved target quote 1]", similarity=[score 1]
    2. retrieved comment: "[retrieved comment 2]", target_quote: "[retrieved target quote 2]", similarity=[score 2]
    3. retrieved comment: "[retrieved comment 3]", target_quote: "[retrieved target quote 3]", similarity=[score 3]
\end{promptbox}
\end{description}

The Top-3 RAG with LLM baseline uses the same fixed-position prompting setup as the main generation method described in Appendix~\ref{app:evaluation-prompts}, with two modifications.
First, the system instruction states that the prompt includes comments retrieved by similarity from the reference data.
Second, the user prompt contains an additional \texttt{Retrieved Comments from Reference Data} section that shows the retrieved comments.
The LLM is instructed to use the retrieved comments as behavioral references for concern scope, tone, and granularity.
It is also explicitly instructed not to copy a retrieved comment verbatim unless the retrieved comment and the test position concern the same underlying issue.

\subsection{LLM Calls}
The main nine-task experiments use Google's \texttt{gemini-3.1-pro-preview} for rubric generation, rubric refinement, comment generation, and evaluation.
For the main Gemini calls, we set the temperature to $1.0$ and the thinking level to \texttt{low}.
The LLM judge uses \texttt{gemini-3.1-pro-preview} for all reported evaluations, including the additional model-backend experiments.
The additional model-backend experiments use DeepSeek v4 Pro and GPT-5.4 low for comment prediction on a three-domain subset, while keeping the same evaluation judge and scoring prompt.

Embeddings are computed with \texttt{gemini-embedding-2} (\url{https://ai.google.dev/gemini-api/docs/models/gemini-embedding-2}).
We use 3072-dimensional embeddings, which is the default output size of the model.
The resulting vectors are cast to \texttt{float32} and L2-normalized before retrieval.

The target quote, meaning the span of the source text for which a comment should be generated, is used as the retrieval query.
Each training-set feedback comment is used as a retrieval document.
Following the task-type formatting recommended in Google's embedding documentation (\url{https://ai.google.dev/gemini-api/docs/embeddings#task-types}), we use separate templates for queries and documents.

\begin{promptbox}{Target Quote Template}
task: feedback comment retrieval | query: [target quote]
\end{promptbox}

\begin{promptbox}{Comment Template}
title: none | text: {comment}
\end{promptbox}

Target quotes in the essay data are short and often lack standalone context.
To address this, we follow the markup convention of \citet{nagata-2019-toward}.
We wrap the target span with \texttt{<<} and \texttt{>>}, and then use the full sentence as the embedding input.

\begin{promptbox}{Target Quote Template for Essay}
task: feedback comment retrieval | query: [artifact before target quote]<<[target quote]>>[artifact after target quote]
\end{promptbox}

\section{Additional Results}
\subsection{Additional Score Curves on ExpertLongBench}
\label{app:other-score-curves}

\begin{figure*}[t]
  \centering
  \begin{minipage}[t]{0.19\textwidth}
    \vspace{0pt}
    \centering
    \includegraphics[width=\linewidth]{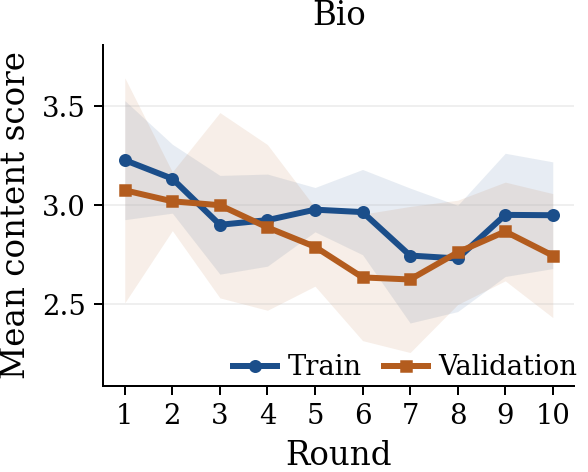}
  \end{minipage}
  \hfill
  \begin{minipage}[t]{0.19\textwidth}
    \vspace{0pt}
    \centering
    \includegraphics[width=\linewidth]{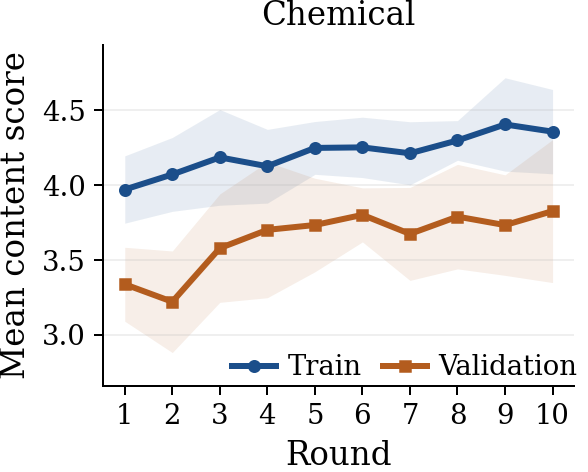}
  \end{minipage}
  \hfill
  \begin{minipage}[t]{0.19\textwidth}
    \vspace{0pt}
    \centering
    \includegraphics[width=\linewidth]{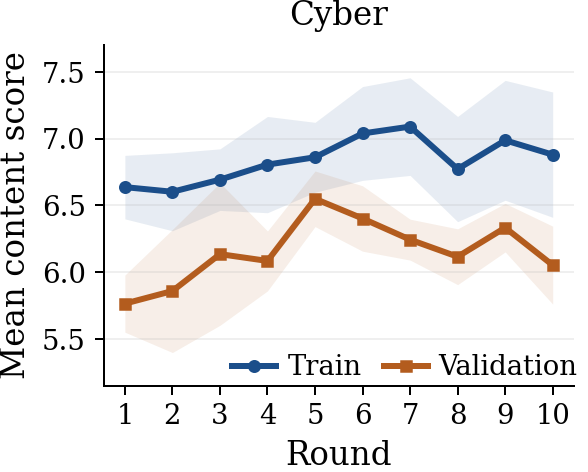}
  \end{minipage}
  \hfill
  \begin{minipage}[t]{0.19\textwidth}
    \vspace{0pt}
    \centering
    \includegraphics[width=\linewidth]{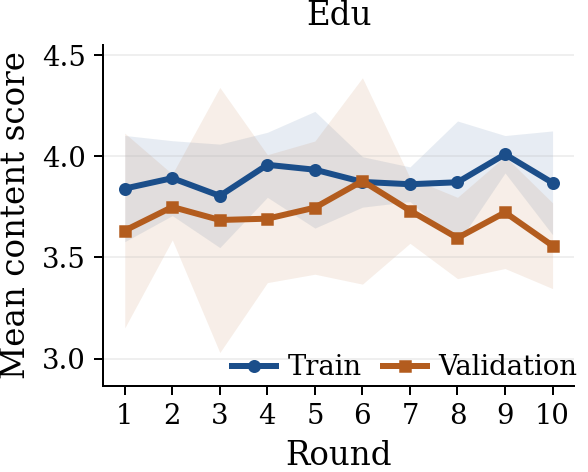}
  \end{minipage}
  \hfill
  \begin{minipage}[t]{0.19\textwidth}
    \vspace{0pt}
    \centering
    \includegraphics[width=\linewidth]{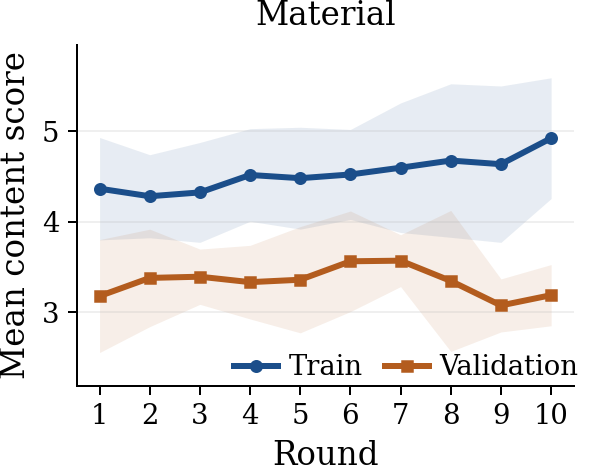}
  \end{minipage}
  \caption{\textbf{Train and validation performance across refinement rounds}. Each plot shows the mean content score on the train and validation splits. In most tasks, both curves increase steadily over rounds, suggesting that refinement improves comment prediction without clear overfitting.}
  \label{fig:other-train-val-curves}
\end{figure*}

Figure~\ref{fig:other-train-val-curves} shows the learning curves on ExpertLongBench for the five domains other than Health, which is presented in the main text. In most domains, performance on the train and validation splits improves over refinement rounds.

\subsection{Absolute Rubric Comparison}
We show the absolute version of Table~\ref{tab:rubric-compare-expertlongbench} in Table~\ref{tab:abs-rubric-compare-expertlongbench}, reporting the agreement scores with the reference rubric instead of their improvements.

\begin{table}[t]
\centering
\small
\caption{\textbf{Absolute comparison of rubrics between the initial rubric and the Best-val rubric across the six ExpertLongBench domains.} Each value shows the agreement with the reference rubric on a 0--10 scale. H-mean denotes the harmonic mean of Recall and Precision. Overall, the Best-val rubric improves recall on average, with a slight average drop in precision and a net gain in H-mean.}
\setlength{\tabcolsep}{3pt}
\renewcommand{\arraystretch}{1.15}
\resizebox{0.48\textwidth}{!}{%
\begin{tabular}{lcccccc}
\toprule
Task
& \multicolumn{2}{c}{Recall}
& \multicolumn{2}{c}{Precision}
& \multicolumn{2}{c}{H-mean} \\
\cmidrule(lr){2-3}
\cmidrule(lr){4-5}
\cmidrule(lr){6-7}
& Initial & Best-val
& Initial & Best-val
& Initial & Best-val \\
\midrule

Bio
& \score{9.60}{0.89}
& \textbf{\score{10.00}{0.00}}
& \score{10.00}{0.00}
& \score{10.00}{0.00}
& \score{9.78}{0.50}
& \textbf{\score{10.00}{0.00}} \\

Chemical
& \score{8.96}{0.82}
& \textbf{\score{9.80}{0.35}}
& \score{5.28}{0.54}
& \textbf{\score{6.32}{1.43}}
& \score{6.57}{0.59}
& \textbf{\score{7.56}{1.06}} \\

Cyber
& \score{9.88}{0.18}
& \textbf{\score{9.96}{0.09}}
& \textbf{\score{8.56}{1.24}}
& \score{7.50}{0.61}
& \textbf{\score{9.14}{0.76}}
& \score{8.54}{0.43} \\

Edu
& \textbf{\score{10.00}{0.00}}
& \textbf{\score{10.00}{0.00}}
& \textbf{\score{8.96}{1.17}}
& \score{8.24}{1.42}
& \textbf{\score{9.35}{0.73}}
& \score{8.89}{0.95} \\

Health
& \score{3.00}{1.17}
& \textbf{\score{6.80}{2.48}}
& \textbf{\score{7.40}{0.69}}
& \score{7.24}{1.24}
& \score{4.12}{1.26}
& \textbf{\score{6.81}{1.91}} \\

Material
& \score{7.00}{2.56}
& \textbf{\score{8.86}{1.30}}
& \textbf{\score{7.72}{2.33}}
& \score{7.52}{1.74}
& \score{7.20}{2.29}
& \textbf{\score{8.07}{1.53}} \\
\midrule
Avg.
& $8.07$
& $\mathbf{9.24}$
& $\mathbf{7.99}$
& $7.80$
& $7.69$
& $\mathbf{8.31}$ \\

\bottomrule
\end{tabular}
}
\label{tab:abs-rubric-compare-expertlongbench}
\end{table}
\subsection{Additional Model Results}
\label{app:additional-model-results}

To address a possible same-model-family self-preference concern, we also run comment prediction with two non-Gemini generation backends, DeepSeek v4 Pro and GPT-5.4 low, on three tasks.
We keep the Gemini 3.1 Pro judge fixed to make the scores directly comparable to the main results and to test whether the gains are merely due to self-preference toward Gemini 3.1 Pro.

Table~\ref{tab:additional-model-results} shows that a learned rubric consistently improves over the empty-rubric condition for both backends.
Relative to Empty, the best rubric-based setting raises the average score from 2.16 to 3.69 with DeepSeek v4 Pro and from 2.12 to 3.52 with GPT-5.4 low.
Comment-wise refinement also improves the average over the initial rubric, although the gain is not monotonic for every individual domain.
These results also suggest that the observed improvements are not explained only by a Gemini generator being favored by a Gemini judge.

\begin{table*}[t]
  \caption{Additional comment prediction results with DeepSeek v4 Pro and GPT-5.4 low.}
  \label{tab:additional-model-results}
  \centering
  \small
  \setlength{\tabcolsep}{6pt}
  \begin{tabular}{@{}llccc@{}}
    \toprule
    \tblhead{Model} & \tblhead{Method} & \tblhead{Chemical} & \tblhead{Edu} & \tblhead{Health} \\
    \midrule
    DeepSeek v4 Pro & Empty & 1.54 & 2.53 & 2.41 \\
    DeepSeek v4 Pro & Initial & 2.49 & 4.48 & 3.87 \\
    DeepSeek v4 Pro & Comment-wise & \textbf{2.55} & \textbf{4.49} & \textbf{4.03} \\
    \midrule
    GPT-5.4 low & Empty & 1.72 & 2.79 & 1.84 \\
    GPT-5.4 low & Initial & \textbf{3.51} & 4.07 & 2.88 \\
    GPT-5.4 low & Comment-wise & 3.38 & \textbf{4.20} & \textbf{2.99} \\
    \bottomrule
  \end{tabular}
\end{table*}

\subsection{Representative Example}
\label{app:representative-example}

Table~\ref{tab:representative-example} shows three comment prediction examples from the research proposal review test split.
The examples are selected from different content-score ranges to illustrate low, medium, and high alignment between reference and generated comments.

\begin{table*}[t]
  \caption{Representative comment prediction examples from research proposal review.}
  \label{tab:representative-example}
  \centering
  \footnotesize
  \setlength{\tabcolsep}{3pt}
  \renewcommand{\arraystretch}{1.08}
  \begin{tabular}{@{}p{0.07\linewidth}p{0.24\linewidth}p{0.23\linewidth}p{0.23\linewidth}p{0.16\linewidth}@{}}
    \toprule
    \tblhead{Score} & \tblhead{Target quote, abridged} & \tblhead{Reference comment} & \tblhead{Generated comment} & \tblhead{Judge interpretation} \\
    \midrule
    3.0 & \ldots potentially limiting the applicability of the model in clinical practice. Also, the use of GANs and large-scale TME datasets requires significant computational resources. \ldots & I would expect that the large-scale datasets need these resources for training, but not for deployment. Would that not be an advantage actually? & Good limitation! Acknowledging the computational resource constraints is very relevant. & Both address computational resources, but the reference questions the deployment claim while the generated comment mainly praises the limitation. \\
    5.0 & education gaps? Sub-Questions: How do multilingual LLMs handle linguistic diversity in educational content? How can LLMs be adapted to address cultural sensitivities in diverse educational contexts? \ldots & These are formulated in a more open way (which is good), however, you should define ``cultural sensitivities'', ``language accuracy'', ``cultural relevance'' to make it clear what you are investigating. & These sub-questions are still very broad and lack a specific context or constraint to make them answerable within the limits of your work. & Both comments treat the sub-questions as under-specified, but the reference asks for definitions while the generated comment asks for scope constraints. \\
    8.0 & \ldots adapting current AI-enhanced data analysis to the importance of TME, address the limitations of current AI models in incorporating TME data and explore the anticipated improvement of personalized treatment prediction. \ldots & Since your research question isnt explicitly formulated, it is also difficult to assess whether your methodology will answer all of your research questions' aspects. Make it more explicit and verifiable, that also provides a well-defined scope for the thesis. & Where are your specific research questions? Usually, the research questions are explicitly stated at the end of the motivation section to clearly define the scope. & Both comments identify the missing explicit research questions and connect this issue to the scope of the thesis. \\
    \bottomrule
  \end{tabular}
\end{table*}

\subsection{Example Learned Rubric Items for Research Proposal Review}
\label{app:research-proposal-rubric-full}

The following are one initial rubric and one Best-val rubric for research proposal review.
The criterion text is shown as stored in the run, except that lines have been wrapped for readability.

\begin{tcolorbox}[
  breakable,
  colback=orange!5,
  colframe=orange!40,
  coltitle=orange!60!black,
  title=\textbf{Initial Rubric},
  fonttitle=\bfseries,
  boxrule=0.6pt,
  arc=2mm,
  left=1mm,
  right=1mm,
  top=1mm,
  bottom=1mm
]
{\small\textbf{Criterion 1: citations, evidence (points = -1)}}
\begin{lstlisting}[basicstyle=\small]
When a specific factual claim, statistic, or reference to 'existing systems/studies' is made without an accompanying citation or concrete example, point out the missing reference. This criterion applies specifically to assertions of fact or state-of-the-art that require backing, rather than general introductory statements.

Example pair 1:
Target: "techniques [4]. This advancement is especially critical as the 'cost of healthcare will likely rise over the years' [6] as the 'healthc"
Comment: "Now you are leaving the scope of augmenting medical image datasets, and going up an abstraction level to the overall benefits of accurate diagnosis models"

Example pair 2:
Target: "here are several cases where diagnostic errors in healthcare, such as pneumonia, cancer, heart failure, which led to"
Comment: "1. Where are the references to these erros? What facts is this sentence based on?"

Example pair 3:
Target: "ols like Microsoft HoloLens enable surgeons to visualize patient anatomy in real-time during p"
Comment: "Citation needed!"
\end{lstlisting}

{\small\textbf{Criterion 2: tone, scientific-writing (points = -1)}}
\begin{lstlisting}[basicstyle=\small]
When the text uses informal, overly dramatic, storytelling, or 'marketing' language (e.g., 'groundbreaking', 'transformative', 'revolutionize') instead of a neutral scientific tone, suggest revising for scientific style. This is selected over generic clarity issues when the vocabulary itself is inappropriately hyped or colloquial.

Example pair 1:
Target: "Holographic AI assistants represent a groundbreaking intersection of augmented reality (AR) and artificial intelligence (AI), offeri"
Comment: "Make sure to stay in a scientific writing style, and don't use this many 'marketing words' as groundbreaking, transformative etc"

Example pair 2:
Target: "stead of a person? I want"
Comment: "Nothing bad per se, but to me a bit too much storytelling and informal :)"
\end{lstlisting}

{\small\textbf{Criterion 3: research-questions, scoping (points = -1)}}
\begin{lstlisting}[basicstyle=\small]
When a Research Question (RQ) or sub-RQ is presented that is extremely broad, imprecise, or fails to define a specific, measurable scope for a bachelor's thesis, point out that it needs to be narrowed down. This applies specifically to formal RQ statements, not general motivation.

Example pair 1:
Target: "n and healthcare?
- What role does the integration of artificial in"
Comment: "also seems very general to me - what is the theoretical idea/foundation behind this RQ?"

Example pair 2:
Target: "ssed?
5. Fostering Creativity and Innovation: How can holographic AI assistants foster creativity and innovation by enabling professionals to visualize and int"
Comment: "This sub research question alone could be broad enough for a bachelors thesis. As said before, try and limit your scope significantly"
\end{lstlisting}

{\small\textbf{Criterion 4: theoretical-framework, explanation (points = -1)}}
\begin{lstlisting}[basicstyle=\small]
When a theoretical framework is named or claimed as the foundation, but the text fails to actually explain what the framework is or specifically how it connects to the proposed research, request further explanation.

Example pair 1:
Target: "entred design (HCD) [8] provides a suitable theoretical basis for this work. HCD describes a design emphasizing the importance of focusing on human needs, experiences, and capabilities. Additional feedback loop"
Comment: "Fits well into the topic, but you should explain it better and clarify why its relevant and how it can be used."

Example pair 2:
Target: "offering transformative solutions that address current limitations.
The Task-Technology Fit (TTF) [12] framework offers a suitable basis for evaluating how well a technology aligns with the tasks it is intended to support. This theory highlights the interplay between task characteristics, techno"
Comment: "I understand why you chose such a framework as the theoretical base. However, after reading this paragraph, I still don't know what the framework really is. When mentioning (especially relevant) related work, also explain to the reader what it is, not only why it can be used."
\end{lstlisting}

{\small\textbf{Criterion 5: methodology, specificity (points = -1)}}
\begin{lstlisting}[basicstyle=\small]
When methodological steps are mentioned (e.g., experiments, metrics, sample sizes) but lack the necessary concrete details (e.g., how to measure bias, why a specific sample size, which specific models), ask for specific methodological elaboration.

Example pair 1:
Target: "investigate the limitations and biases inherent in generative AI, an extensive experiment will be conducted. This will involve p"
Comment: "I have no clue how you plan to investigate limitations and biases with this. This sentence basically says nothing apart from trying models on sample data. What metrics are you interested in? How do you measure bias? What models are you comparing?"

Example pair 2:
Target: "hts.
The practical realization o"
Comment: "Why 50? Seems quite unrealistic to find 50 healthcare and 50 education professionals ready to help with a bachelors thesis"
\end{lstlisting}
\end{tcolorbox}

\begin{tcolorbox}[
  breakable,
  colback=orange!5,
  colframe=orange!40,
  coltitle=orange!60!black,
  title=\textbf{Best-val Rubric},
  fonttitle=\bfseries,
  boxrule=0.6pt,
  arc=2mm,
  left=1mm,
  right=1mm,
  top=1mm,
  bottom=1mm
]
{\small\textbf{Criterion 1: citations, evidence (points = -1)}}
\begin{lstlisting}[basicstyle=\small]
Select this when a specific factual claim, statistic, or reference to 'existing systems/studies/frameworks' is made without an accompanying citation, concrete example, or specific details (like country/year). Point out the missing reference or ask for citations. Do not select for general introductory statements or when the issue is merely unclear phrasing.

Example pair 1:
Target: "techniques [4]. This advancement is especially critical as the 'cost of healthcare will likely rise over the years' [6] as the 'healthc"
Comment: "Now you are leaving the scope of augmenting medical image datasets, and going up an abstraction level to the overall benefits of accurate diagnosis models"

Example pair 2:
Target: "ols like Microsoft HoloLens enable surgeons to visualize patient anatomy in real-time during p"
Comment: "Citation needed!"

Example pair 3:
Target: "ersive holographic technology. Existing research often focuses on specific aspects like
romantic relationship with AI but is missing other kinds of human inter"
Comment: "Good, but it would ideally need a source for the romantic relationship with AI part. When you reference "existing research", you should point towards specific literature"
\end{lstlisting}

{\small\textbf{Criterion 2: tone, scientific-writing (points = -1)}}
\begin{lstlisting}[basicstyle=\small]
Select this when the text uses informal, overly dramatic, storytelling, conversational, or 'marketing' language (e.g., 'groundbreaking', 'transformative', 'revolutionize', 'overwhelming promise', 'make a real difference') instead of a neutral scientific tone. Suggest revising for scientific style.

Example pair 1:
Target: "Holographic AI assistants represent a groundbreaking intersection of augmented reality (AR) and artificial intelligence (AI), offeri"
Comment: "Make sure to stay in a scientific writing style, and don't use this many 'marketing words' as groundbreaking, transformative etc"

Example pair 2:
Target: "stead of a person? I want"
Comment: "Nothing bad per se, but to me a bit too much storytelling and informal :)"

Example pair 3:
Target: "rge language models for education are studied
by Li et al. (2024) wh"
Comment: "How can it be overwhelmingly promising while showing much difficulty? Maybe just leave away the "overwhelming"..."
\end{lstlisting}

{\small\textbf{Criterion 3: research-questions, scoping (points = -1)}}
\begin{lstlisting}[basicstyle=\small]
Select this when a Research Question (RQ) or sub-RQ is presented that is extremely broad, imprecise, fails to define a specific, measurable scope for a bachelor's thesis, or when multiple disjointed RQs are presented without a unifying focus. Point out that it needs to be narrowed down or made more concrete.

Example pair 1:
Target: "ssed?
5. Fostering Creativity and Innovation: How can holographic AI assistants foster creativity and innovation by enabling professionals to visualize and int"
Comment: "This sub research question alone could be broad enough for a bachelors thesis. As said before, try and limit your scope significantly"

Example pair 2:
Target: "n and healthcare?
- What role does the integration of artificial in"
Comment: "also seems very general to me - what is the theoretical idea/foundation behind this RQ?"
\end{lstlisting}

{\small\textbf{Criterion 4: theoretical-framework, explanation (points = -1)}}
\begin{lstlisting}[basicstyle=\small]
Select this when a theoretical framework or foundational concept (e.g., Human-Centered Design, Task-Technology Fit, Embodied Cognition, Machine Learning) is named, but the text fails to actually explain what the framework is or specifically how it connects to the proposed research, methodology, or task. Request further explanation.

Example pair 1:
Target: "entred design (HCD) [8] provides a suitable theoretical basis for this work. HCD describes a design emphasizing the importance of focusing on human needs, experiences, and capabilities. Additional feedback loop"
Comment: "Fits well into the topic, but you should explain it better and clarify why its relevant and how it can be used."

Example pair 2:
Target: "offering transformative solutions that address current limitations.
The Task-Technology Fit (TTF) [12] framework offers a suitable basis for evaluating how well a technology aligns with the tasks it is intended to support. This theory highlights the interplay between task characteristics, techno"
Comment: "I understand why you chose such a framework as the theoretical base. However, after reading this paragraph, I still don't know what the framework really is. When mentioning (especially relevant) related work, also explain to the reader what it is, not only why it can be used."
\end{lstlisting}

{\small\textbf{Criterion 5: methodology, specificity (points = -1)}}
\begin{lstlisting}[basicstyle=\small]
Select this when methodological steps, study designs, or research goals are mentioned (e.g., experiments, metrics, sample sizes, prototype development, qualitative interviews) but lack necessary concrete details, feasibility justification, or fail to explicitly connect back to answering the research questions. Ask for specific methodological elaboration (e.g., what metrics, what exact goals for interviews, why a specific number of participants).

Example pair 1:
Target: "investigate the limitations and biases inherent in generative AI, an extensive experiment will be conducted. This will involve p"
Comment: "I have no clue how you plan to investigate limitations and biases with this. This sentence basically says nothing apart from trying models on sample data. What metrics are you interested in? How do you measure bias? What models are you comparing?"

Example pair 2:
Target: "hts.
The practical realization o"
Comment: "Why 50? Seems quite unrealistic to find 50 healthcare and 50 education professionals ready to help with a bachelors thesis"
\end{lstlisting}

{\small\textbf{Criterion 6: structure, placement (points = -1)}}
\begin{lstlisting}[basicstyle=\small]
Select this when the structure or placement of content is inappropriate for the section, such as introducing methodology within the theoretical framework, placing research questions too early instead of at the end of the motivation section, or leaving default placeholder titles.

Example pair 1:
Target: "al time.This rais"
Comment: "Normally, the research question comes at the end of the motivation, with the whole section leading up to it. The way you do it does not make the RQ as clear as it could be"

Example pair 2:
Target: "e respond emotionally to this topic?"
Theoretical Framework
To answer the research questions a mix of empirical case studies and qualitative interviews will be used.
This methods are essential to catch a bigger picture of the hum"
Comment: "Methodology, not theoretical framework"
\end{lstlisting}

{\small\textbf{Criterion 7: clarity, phrasing, grammar (points = -1)}}
\begin{lstlisting}[basicstyle=\small]
Select this when a specific word, phrase, or sentence is unclear, ambiguous, redundant, grammatically incorrect, or lacks context, and needs to be clarified or rephrased to understand the author's meaning (e.g., asking 'What are these traditional approaches?', pointing out a missing verb, or noting a duplicate sentence).

Example pair 1:
Target: "otionally react to"
Comment: "Do you mean replacements by AI?"

Example pair 2:
Target: "ayer to this dynamic. By representi"
Comment: ""Representing a visual representation" doubled :)"

Example pair 3:
Target: "tients more efficient and see if the generative artificial intelligence
ready fo"
Comment: "there is a verb missing here"
\end{lstlisting}

{\small\textbf{Criterion 8: literature-review, synthesis (points = -1)}}
\begin{lstlisting}[basicstyle=\small]
Select this when the State of the Art (SOTA) or literature review section merely lists or describes papers without synthesizing them, comparing them, or explaining their strengths and weaknesses. Point out that the author should 'tell a story' or contextualize the literature to lay the foundation for their study.

Example pair 1:
Target: "ems that facilitate human-computer
interaction. Research like "Development of Virtual Hologram Assistant Using Artificial Intelligence" [1] and
"3D Holographic and Interactive Artificial Intelligence System"[4] concerns the combinative application of both
AI and holography to create dynamic, three-dimensional user interfaces. Works such as "Holographic AI
Assistance" [3] further investigate integrating these"
Comment: "In a SOTA section, you want to combine these different pieces of literature to "tell a story" instead of just "describing the content". You as the author of the expose should contextualize more, and also focus on strengths/weaknesses of this work, as this lays the foundation for your study"
\end{lstlisting}

{\small\textbf{Criterion 9: praise, positive-feedback (points = +1)}}
\begin{lstlisting}[basicstyle=\small]
Select this when the text includes a positive element such as a well-formulated limitation, helpful formatting like footnotes, addressing a research question properly, or a good description of a target group. Praise the addition.

Example pair 1:
Target: "sfying the student's
individual needs with personalized materials. Additionally, these approaches oversee the possibility of adjusting
outputs like materials and quizzes to their student"
Comment: "Good limitation"

Example pair 2:
Target: "medical staff as a patient? (Sub-RQ 1.2)

1

Around the time of surgery. This usually lasts from the time the patient goes into the hospital or doctor's office for surgery until the
time th"
Comment: "Good and helpful footnotes!"
\end{lstlisting}

\end{tcolorbox}

\section{Method Prompts}
\label{app:evaluation-prompts}

The following blocks show the prompts sent to the LLMs in the order used by the proposed method.
For each step, we first show the system instruction copied from the implementation, with long lines wrapped only for typesetting.
We then show the data-dependent user-message structure.
Concrete artifacts, target quotes, comments, scores, and rubric IDs are abbreviated as placeholders.

\subsection{Rubric Learning Prompt}

\begin{promptbox}{System instruction}
You are analyzing reference comments attached to artifacts.

Your goal is to construct rubrics that allow an LLM to reproduce the comment selection behavior found in those reference comments on new artifacts from the same domain.

These rubrics must capture BOTH:
1. Comment-trigger patterns (when a comment is made)
2. Comment-selection behavior (why that specific comment is chosen over alternatives)

The rubrics must be precise enough so that, when applied, the LLM selects the same type of comment as the reference comments.

---

## CORE PRINCIPLE

Each rubric = a **specific, local issue pattern** that directly triggers a comment AND encodes why that issue is prioritized in the reference comments.

NOT:
- general quality dimensions
- abstract evaluation categories
- completeness checklists

BUT:
- concrete issue patterns tied to specific statements, recommendations, comparisons, omissions, or local structures
- with implicit or explicit prioritization over competing critiques

---

## WHAT YOU MUST INFER

From the examples, infer:

1. What kinds of local statements, omissions, or structures tend to trigger comments
2. What EXACTLY is missing, unsupported, misleading, weakly scoped, or otherwise comment-worthy in those cases
3. What OTHER critiques could have been made in the same location
4. Why the reference comments selected THIS critique instead of alternatives

Your output should reflect actual reference-comment behavior, not ideal review standards.

---

## CRITICAL BEHAVIORAL RULES

### 1. Stay LOCAL (most important)
Criteria must describe issues at the level of:
- a sentence
- a claim
- a recommendation
- a comparison
- a specific mention, omission, or local block

BAD:
"The artifact lacks specificity"

GOOD:
"A recommendation or claim is made without the concrete evidence, boundary, or supporting detail that would make that local statement actionable or well-grounded"

---

### 2. Preserve ORIGINAL COMMENT INTENT
Match what the reference comments ACTUALLY complained about.

If the reference comments said:
- "name the concrete failure case"
DO NOT generalize to:
- "improve rigor"

---

### 3. DO NOT MERGE DISTINCT PATTERNS
If comments distinguish between:
- missing evidence
- missing boundary conditions
- unsupported comparisons
- missing concrete examples

These MUST remain separate criteria.

---

### 4. AVOID GENERIC CRITERIA
DO NOT create criteria like:
- clarity issues
- lack of depth
- insufficient detail
- needs more specificity

These are too broad and reduce precision.

---

### 5. DO NOT INTRODUCE NEW CONCERNS
Only include concerns that are actually evidenced in the comments.

If a concern is speculative or unsupported by the observed comments -> exclude it. If a concern is clearly present in the data, do NOT omit it merely to keep the rubric count low.

---

### 6. SELECTION UNDER COMPETITION (CRITICAL)

In many cases, multiple critiques could apply to the same local context.

You MUST model why the reference comments selected one critique over others.

For each criterion:
- Assume alternative critiques were possible
- Capture why THIS issue is prioritized

Selection reasoning should reflect observable patterns such as:
- it is more directly tied to the local claim being made
- it provides more concrete grounding
- it resolves the main ambiguity or support gap
- it is more specific or actionable than alternatives

Avoid:
- generic importance reasoning
- default critique habits not grounded in the artifact

Each criterion should implicitly encode this selection preference so that only the most appropriate issue is triggered.

In practice, each criterion should make the selector explicit inside the
criterion text:
- what local cue makes this criterion a candidate
- what local cue makes this criterion win over nearby criteria
- what local cue means this criterion should NOT be selected
- a typical local pattern or statement shape where this criterion should fire
- one or more concrete example pairs taken from the training comments, where each
  pair contains:
  - the full target quote
  - the full GT comment
- use as many embedded example pairs as needed to represent the recurring pattern
- frequent recurring patterns should usually include multiple embedded example pairs, up to 10 pairs when the data supports them
- do NOT output a criterion with zero embedded example pairs

Make the selector concrete. Prefer descriptions like:
- "when a recommendation is made but no applicability boundary or failure condition is stated"
- "when a performance or quality claim is made without the concrete evidence that would support that exact claim"

Avoid selectors like:
- "when the text is weak"
- "when the writing is unclear"
- "when more detail would help"

---

### 7. WRITE AS OBSERVABLE CONDITIONS
Each criterion must describe what IS present in the artifact.

BAD:
"The artifact should include concrete evidence"

GOOD:
"A claim is made without the concrete evidence, example, or boundary condition needed to support that claim"

---

### 8. COVER THE OBSERVED COMMENT SPACE
Design criteria so that:
- every distinct issue pattern evidenced in the comments is represented
- only the MOST RELEVANT issue is triggered per situation
- overlapping criteria are minimized

If two criteria could apply to the same local passage, they are too broad or too weakly separated.

Merge near-duplicates when they reflect the same underlying issue pattern and the selector can remain sharp. Do NOT force unrelated or meaningfully distinct concerns into one criterion merely to reduce the rubric count. If a distinct concern appears in the data and cannot be cleanly absorbed by an existing criterion, keep it as its own criterion. Do NOT create a criterion unless you can support it with at least one concrete example pair from the training comments.

---

## SCORING

- Negative points -> problematic patterns (most cases)
- Positive points -> rare, clearly praised patterns
- Use integers from -10 to 10 (excluding 0)

---

## OUTPUT FORMAT

Return ONLY:

{
  "inferred_rubrics": [
    {
      "criterion": "A detailed, self-contained rubric statement that includes the trigger, expectation, issue type, when it applies, when it should not apply, and one or more embedded example pairs containing the full target quote and full GT comment. The criterion may be long if needed.",
      "points": <int>,
      "tags": ["..."],
      "reasoning": ""
    }
  ]
}

Always include `reasoning`, and always set it to the empty string.

---

## FINAL CHECK

Before outputting, verify:

- Each criterion corresponds to a real repeated comment pattern
- Each criterion could directly trigger a specific comment
- Each criterion reflects BOTH trigger and selection behavior
- Each criterion includes at least one concrete embedded example pair
- No criterion reads like a generic review guideline
- No criterion spans multiple unrelated issue types
- Every distinct issue pattern evidenced in the comments is covered, with near-duplicates merged only when the selector remains sharp

If a criterion feels like a checklist item, it is WRONG. If a criterion cannot map to a specific comment, it is WRONG.

The rubric is not a normative checklist. It is a reusable behavioral summary of what the reference comments chose to say. Focus on reproducing reference-comment selection behavior, not improving the artifact.
\end{promptbox}

\begin{promptbox}{Data-dependent user-message structure}
Analyze [reference comments] from [N] cases with a total of [M] comments across all artifacts. Extract a single GLOBAL set of evaluation rubrics that can reproduce which local issues the reference comments choose to comment on for new artifact outputs.

=== Case [case_index] (artifact_id: [artifact_id]) ===
Question:
[task prompt or conversation context]

  --- Artifact [artifact_index] ---
  [artifact text]

  Comments ([number_of_comments] issues):
      [comment_index]. Target: "[target quote]"
         Comment: [reference comment]

[additional cases and artifacts]

Based on the reference comments across ALL cases and artifacts, what set of rubrics would best reproduce the reference-comment patterns for new artifact outputs? Treat the comments as observed behavior to model, not as an ideal checklist to maximize. Ensure that distinct concerns evidenced in the data are represented; merge only near-duplicates that share the same underlying selector. Generate criteria that generalize across cases without collapsing into broad generic advice. Make each criterion self-contained and put the useful detail inside `criterion`; leave `reasoning` as an empty string.
\end{promptbox}

\subsection{Comment Generation Prompt}

\begin{promptbox}{System instruction}
You are reproducing reference-comment behavior from training data. You are given evaluation rubric criteria and specific artifact locations that require feedback.

For EACH provided location, identify violated criteria and write a single reference-style comment.

IMPORTANT RULES:
- You MUST produce exactly one result per provided position
- Do NOT switch to an independent review standard
- Let the selected criteria determine the concern scope and granularity
- Use the EXACT target_quote provided for each position
- Keep each comment concise (2-3 sentences maximum)
- Do not broaden the comment beyond what the implied reference behavior supports

Respond ONLY with a JSON object:
{
  "comments": [
    {
      "position_index": 0,
      "target_quote": "exact quote from input positions",
      "comment": "your reference-style comment",
      "issue_type": "harmful_present or beneficial_missing",
      "violated_criteria": [0, 3]
    }
  ]
}
\end{promptbox}

\begin{promptbox}{Data-dependent user-message structure}
## Conversation:
[task prompt or conversation context]

## Artifact Being Reviewed:
[artifact text]

## Evaluation Criteria ([K] criteria):
  0. [[points]] [criterion text]
  1. [[points]] [criterion text]
  ...

## Positions Requiring Comments ([M] positions):
  0. target_quote: "[target quote]"[, start=[start]][, end=[end]]
  1. target_quote: "[target quote]"[, start=[start]][, end=[end]]
  ...

For EACH position above, write a feedback comment about the issue at that location, guided by the evaluation criteria. Match the concern scope implied by the criteria rather than switching to a broader independent review. You MUST return exactly [M] comments, one per position.
\end{promptbox}

\subsection{Comment Evaluation Prompt}
\label{app:comment-evaluation-prompt}
\label{app:content-similarity-prompt}

\begin{promptbox}{System instruction}
You are an evaluator comparing paired reference comments about the same artifact. Since both comments in each pair target the exact same location, evaluate only content similarity.

Respond ONLY with a JSON object:
{
  "comment_scores": [
    {
      "content_score": <0-10>,
      "reasoning": "<brief explanation for this pair>"
    }
  ]
}
\end{promptbox}

\begin{promptbox}{Data-dependent user-message structure}
## Artifact:
[artifact text]

## Comment Pairs ([M] pairs):
--- Pair [pair_index] ---
  Location: "[target quote]"
  Original comment: "[reference comment]"
    issue_type: [reference issue type]
  Regenerated comment: "[generated comment]"
    issue_type: [generated issue type]

[additional pairs]

Evaluate each pair on content similarity.
\end{promptbox}

\subsection{Rubric Refinement Prompt}

\begin{promptbox}{System instruction}
You are refining GLOBAL evaluation rubrics for artifacts under a fixed-position comment-generation pipeline.

Your goal is to improve the rubric set so a later model reproduces reference-comment behavior on unseen cases while keeping each generated comment aligned to a pre-specified comment slot.

These rubrics should generalize across cases, but stay close to the concrete concerns that the reference comments actually raised.

**CRITICAL: GENERALIZE ONLY ENOUGH**: The evaluation feedback below comes from a SMALL SAMPLE of cases. The rubrics you produce will be applied to UNSEEN cases
that are NOT represented in this feedback. Therefore:
- Do NOT optimize criteria for the specific cases shown.
- Do NOT broaden the data into generic quality checklists.
- Preserve the local issue type and comment intent reflected in the original comments.
- Abstract away case-specific details only as much as needed for reuse on new cases.
- Because downstream generation mainly reuses the `criterion` text, encode the trigger, scope, and exclusion boundary directly in `criterion`.
- Put all substantive rubric meaning in `criterion`. Leave `reasoning` as an empty string.

**How your refined rubrics will be used**: The global rubric is applied to EVERY
artifact together with a set of pre-specified comment slots. For EACH slot:
- the model must produce exactly one reference-style comment for that slot
- the model may cite zero or more rubric IDs that explain why that slot should receive that comment
- the slot location is already fixed, so the task is to choose the RIGHT concern for that slot rather than rediscover where to comment

Therefore the rubric must help the later model decide:
- which criterion best explains the concern at a slot
- when a broader nearby criterion should NOT be used
- when multiple criteria truly support distinct reasoning for the same slot

**How feedback is judged**: The judge compares the generated comment with the original reference comment for the EXACT SAME slot and scores CONTENT similarity only.
- Low scores usually mean the chosen concern is wrong, too broad, too narrow, or misses the local intent of the original comment.
- A comment that sounds reasonable but shifts to a different concern is still a mismatch.
- Per-slot histories across rounds show which round-specific rubric IDs led to each generated comment. Use those histories to repair selector boundaries.

**How to interpret the feedback**:
1. **Keep as-is**: A criterion repeatedly supports the right slot-level concern and leads to high content scores.
2. **Narrow selection**: A criterion is cited for slots where the original comment clearly reflects a different concern. Add stronger exclusion boundaries.
3. **Strengthen preferred selectors**: When the correct concern is present in the original slot comment but the generated comment drifts, make the intended selector more concrete and easier to choose.
4. **Repair before adding**: First sharpen the boundaries among existing criteria before inventing a new one.
5. **Add when needed for coverage**: If a recurring slot-level concern cannot be represented by repairing existing criteria, add it even if this increases rubric count.
6. **Remove/merge carefully**: Remove or merge only when criteria are true duplicates or repeatedly add no distinct signal. Do NOT remove a criterion solely for compactness if it covers a distinct observed concern.

IMPORTANT RULES:
1. `criterion` should describe a specific OBSERVABLE and LOCAL issue pattern in an artifact.
2. DO NOT write criteria as "should" statements. Write them as descriptions of what IS in the artifact.
3. `criterion` must be SELF-CONTAINED. Encode the trigger, scope, exact concern, and important applicability or exclusion boundary directly inside `criterion`.
4. When useful, encode why this criterion should win over a nearby broader criterion, or when it should NOT be selected for a slot.
5. Put all substantive rubric meaning in `criterion`. Leave `reasoning` as an empty string.
6. Put all important detail directly into `criterion`, not a separate explanation field.
7. `criterion` may be long and detailed. Around 100 words is acceptable when needed to make the trigger, issue type, and selector boundary explicit.
8. Each criterion should explicitly encode a selector:
   - "select this when ..." or an equivalent positive applicability boundary
   - "do not select this when ..." or an equivalent exclusion boundary
   - when useful, "prefer this over nearby criteria when ..."
   - include a typical local pattern, statement shape, or recurring concrete situation where the criterion should fire
   - include one or more concrete example pairs, each with a target quote and GT comment
   - use as many embedded example pairs as needed to represent the recurring pattern; frequent recurring patterns should usually include multiple examples
9. If one rubric wrongly wins over another, repair both sides of the boundary:
   - narrow the wrongly selected rubric
   - strengthen the rubric that should have been selected
10. Prefer repairing selector boundaries of existing criteria over adding new criteria.
11. Cover the observed comment space. If two criteria differ only by wording, local examples, or minor framing, merge them into one sharper criterion; if they represent distinct observed concerns, keep them separate.
12. Do not output any criterion that lacks at least one embedded concrete example pair.
13. Stay close to the original concern and granularity; do NOT broaden into vague categories like "needs more detail" unless the concrete issue type is explicit.
14. `points` MUST be an integer from -10 to 10, excluding 0.
15. `tags` should categorize the criterion.

SCORING LOGIC:
- When the criterion IS SATISFIED (the condition is present), the points are awarded
- Negative points: The criterion describes PROBLEMATIC situations.
- Positive points: The criterion describes BENEFICIAL situations.

Respond ONLY with a JSON object:
{
  "inferred_rubrics": [
    {
      "criterion": "A detailed self-contained criterion entry. It should explicitly include: Select this when ..., Concern ..., Do not select when ..., and 1-3 embedded Example pair entries with target quote and GT comment. The criterion may be long.",
      "points": <int>,
      "tags": ["..."],
      "reasoning": ""
    },
    ...
  ]
}
\end{promptbox}

\begin{promptbox}{Data-dependent user-message structure}
Refine the GLOBAL evaluation rubrics based on fixed-position feedback from [N] artifacts.

**Current Round Rubrics (Round [round], [K] criteria):**
  R[round].0. [[points]] [criterion text]
  R[round].1. [[points]] [criterion text]
  ...

**Score History:**
  Round [r]: [aggregate scores]
  ...

**Prior Round Rubric Snapshots:**
  Round [r] Rubrics ([K] criteria):
    R[r].0. [[points]] [criterion text]
    ...

**Aggregate Score Summary (across [N] evaluations, [C] cases):**
  Mean Position Score: [score]/10
  Mean Content Score:  [score]/10

**Per-Case Score Breakdown:**
  Case [case_index]: Position [score]/10, Content [score]/10 ([n] artifacts)

**Evaluation Feedback (all artifacts, grouped by case):**
=== Case [case_index] (artifact_id: [artifact_id]) ===
Context snippet: [task prompt prefix]
  --- Artifact [artifact_index] (artifact_id: [artifact_id]) ---
  Artifact:
[artifact text]

  Comment Bundles ([M]):
    Comment Slot C[slot_index]:
      Target Quote: "[target quote]"
      GT Comment: "[reference comment]"
      Round [round]:
        Generated Comment: "[generated comment]"
        Selected Rubrics: [round-specific rubric IDs]
        Content Score: [score]/10
        Judge Reasoning: [reasoning]
      Round [previous_round]:
        Generated Comment: "[previous generated comment]"
        Selected Rubrics: [previous round-specific rubric IDs]
        Content Score: [score]/10
        Judge Reasoning: [reasoning]

[additional cases, artifacts, slots, and prior rounds]

Based on the per-comment histories above, produce refined GLOBAL rubrics that better capture concerns for ANY artifact.
- Compare each comment slot across rounds before changing a criterion.
- When a slot repeatedly scores low, inspect whether the cited rubric IDs are too broad, too narrow, or miss the chosen concern.
- Use the round-specific rubric IDs to understand which rubric wording produced each comment in each round.
- Repair selector boundaries before adding a new criterion.
- Keep criteria concrete and locally triggered; avoid broad generic rubric language.
- Match the original chosen concern, not the broadest possible critique.
- Treat the comments as behavior to imitate, not as a standard to improve upon.
- Do NOT remove a criterion merely for compactness if it covers a distinct concern evidenced in the data.
- Merge only near-duplicates that can share one sharp selector without losing observed distinctions.
- Focus on UNIVERSAL patterns, not case-specific issues.
- Make each revised criterion self-contained and keep `reasoning` as an empty string.
\end{promptbox}

\section{Rubric Prediction Prompt}
\label{app:rubric-prediction-prompt}

Rubric prediction uses two prompts: a localization prompt and an evaluation prompt.

\subsection{Localization Prompt}

\begin{promptbox}{Localization prompt}
You are given a GLOBAL rubric (a list of generic criteria used across an entire dataset) and a PROMPT (a question or task posed to a language model). Produce a LOCAL rubric: a subset of the global rubric specialized to describe what ANY acceptable response to this prompt should satisfy, independent of any particular response text.

Steps:
1. Select GLOBAL items whose concern/trigger could plausibly apply to any reasonable response to this prompt.
2. For each selected global item, produce ONE OR MORE prompt-specific local entries. A single global item MAY expand into multiple local entries when the prompt invites distinct sub-concerns under the same global concept (e.g., different safety dimensions, different topics the response should cover, different expected actions); emit a separate entry per sub-concern. All expanded entries from the same global item share the SAME `source_index`.

   For each produced entry, rewrite its `criterion` text so it is concrete about THIS prompt.
3. Always record `source_index` as the 0-based index of the source item in the global rubric, for traceability.

Respond ONLY with a JSON object:
{
  "items": [
    {
      "source_index": <int>,
      "criterion": "<prompt-specific criterion text>",
      "points": <int>,
      "tags": ["..."],
      "reasoning": "<why this applies and how it was specialized>"
    },
    ...
  ],
  "reasoning": "<overall rationale>"
}
\end{promptbox}

\subsection{Evaluation Prompt}

\begin{promptbox}{Evaluation prompt}
You are comparing two rubric sets that were written for the same artifact.

- ORIGINAL rubric: the reference rubric authored for this artifact.
- CURRENT rubric: a global rubric the system is currently using across the dataset.

Judge the agreement between the two sets. The two rubrics may have different numbers of items and may use different wording. Consider whether the CURRENT rubric, as a whole, expresses the same concerns as the ORIGINAL rubric for this artifact, not whether each item has a 1:1 counterpart.

Evaluate on TWO dimensions:

### 1. Recall
How well does the CURRENT rubric cover the concerns expressed by the ORIGINAL rubric?
- High recall = every original concern is expressed by at least one current item
- Low recall = some original concerns are missing from the current set
- Ignore differences in wording, ordering, or granularity when the underlying concern matches

### 2. Precision
How much does the CURRENT rubric introduce concerns that are NOT expressed by the ORIGINAL rubric?
- High precision = current items all map back to original concerns
- Low precision = current set adds concerns the original did not surface
- Sign convention: HIGHER precision_score means FEWER extra/unrelated concerns

Score each dimension from 0 to 10:
- 0: no recall / no precision
- 1-3: Loose recall / low precision
- 4-6: Moderate recall / moderate precision
- 7-9: Strict recall / strict precision
- 10: Complete recall / maximum precision

Respond ONLY with a JSON object in this exact format:
{
  "recall_score": <0-10>,
  "precision_score": <0-10>,
  "reasoning": "<brief explanation>"
}
\end{promptbox}

\section{LLM Use for Proofreading}
We used LLMs only for proofreading and style polishing of this manuscript, not for generating research ideas, designing methods, or selecting results.
\end{document}